%% file: main.tex
\documentclass[sigconf]{acmart}

\usepackage{amsmath}

\usepackage[skip=0pt]{caption}
\usepackage{graphicx}
\usepackage{pgfplots}
\pgfplotsset{compat=1.14}
\usepackage{tabularx}
\usepackage{pgf-pie}
\usepackage{caption}
\usepackage{subcaption}
\usepackage{multirow}
\usepackage[inline]{enumitem}

\newcommand{\krimp}{\vspace*{-1mm}}

\AtBeginDocument{%
  \providecommand\BibTeX{{%
    \normalfont B\kern-0.5em{\scshape i\kern-0.25em b}\kern-0.8em\TeX}}}

\acmSubmissionID{66}
\copyrightyear{2020}
\acmYear{2020}
\setcopyright{acmcopyright}
\acmConference[CHIIR '20]{2020 Conference on Human Information Interaction and Retrieval}{March 14--18, 2020}{Vancouver, BC, Canada}
\acmBooktitle{2020 Conference on Human Information Interaction and Retrieval (CHIIR '20), March 14--18, 2020, Vancouver, BC, Canada}
\acmPrice{15.00}
\acmDOI{10.1145/3343413.3377971}
\acmISBN{978-1-4503-6892-6/20/03}

\begin{document}

\title{Conversations with Documents}
\subtitle{An Exploration of Document-Centered Assistance}


\newcommand\blfootnote[1]{%
  \begingroup
  \renewcommand\thefootnote{}\footnote{#1}%
  \addtocounter{footnote}{-1}%
  \endgroup
}
\newcommand{\ruimte}{ \ }

\author{
\mbox{}$\!\!\!$Maartje ter Hoeve$^{1, *}$\ruimte
Robert Sim$^2$\ruimte
Elnaz Nouri$^2$\ruimte
Adam Fourney$^2$\ruimte
Maarten de Rijke$^1$\ruimte
Ryen W. White$^2$
$\!\!\!$\mbox{}
}
\affiliation{%
$^1$University of Amsterdam, Amsterdam, Netherlands\quad
$^2$Microsoft Research, Redmond, WA\\
m.a.terhoeve@uva.nl, rsim@microsoft.com,
elnouri@microsoft.com\\ adamfo@microsoft.com, derijke@uva.nl, ryenw@microsoft.com
}

\renewcommand{\shortauthors}{ter Hoeve et al.}

\begin{abstract}
The role of conversational assistants has become more prevalent in helping people increase their productivity. Document-centered assistance, for example to help an individual quickly review a document, has seen less significant progress, even though it has the potential to tremendously increase a user's productivity. This type of document-centered assistance is the focus of this paper. Our contributions are three-fold: \begin{enumerate*} \item We first present a survey to understand the space of document-centered assistance and the capabilities people expect in this scenario.  \item We investigate the types of queries that users will pose while seeking assistance with documents, and show that document-centered questions form the majority of these queries.
\item We present a set of initial machine learned models that show that (a) we can accurately detect document-centered questions, and (b) we can build reasonably accurate models for answering such questions.  \end{enumerate*} These positive results are encouraging, and suggest that even greater results may be attained with continued study of this interesting and novel problem space. Our findings have implications for the design of intelligent systems to support task completion via natural interactions with documents.
\blfootnote{*Work done while the first author was an intern at Microsoft Research AI.}
\end{abstract}




%% the work being presented. Separate the keywords with commas.%

% fix to get rid of the superscripts in the author names
\let\backupauthors\authors
\renewcommand{\authors}{Maartje ter Hoeve, Robert Sim, Elnaz Nouri, Adam Fourney, Maarten de Rijke, and Ryen W. White}
\maketitle
\renewcommand{\authors}{\backupauthors}

\input{01-introduction.tex}

\input{02-related_work.tex}

\input{03-user_study.tex}
\input{04-QA_data_collection.tex}
\input{05-modelling.tex}
\input{06-discussion_conclusion.tex}

%\krimp
\begin{acks}
We thank Julia Kiseleva for her helpful feedback. This research was supported by Ahold Delhaize, the Association of Universities in the Netherlands (VSNU), the Innovation Center for Artificial Intelligence (ICAI), and the Nationale Politie. All content represents the opinion of the authors, which is not necessarily shared or endorsed by their respective employers and/or sponsors.
\end{acks}

\bibliographystyle{ACM-Reference-Format}
\bibliography{bibliography}

\appendix

\input{07-AppendixA.tex}

\end{document}

%% file: 01-introduction.tex
%% !TEX root = ./main.tex
\section{Introduction}
\label{sec:introduction}

\begin{figure}
	\centering
	\includegraphics[width=0.9\columnwidth]{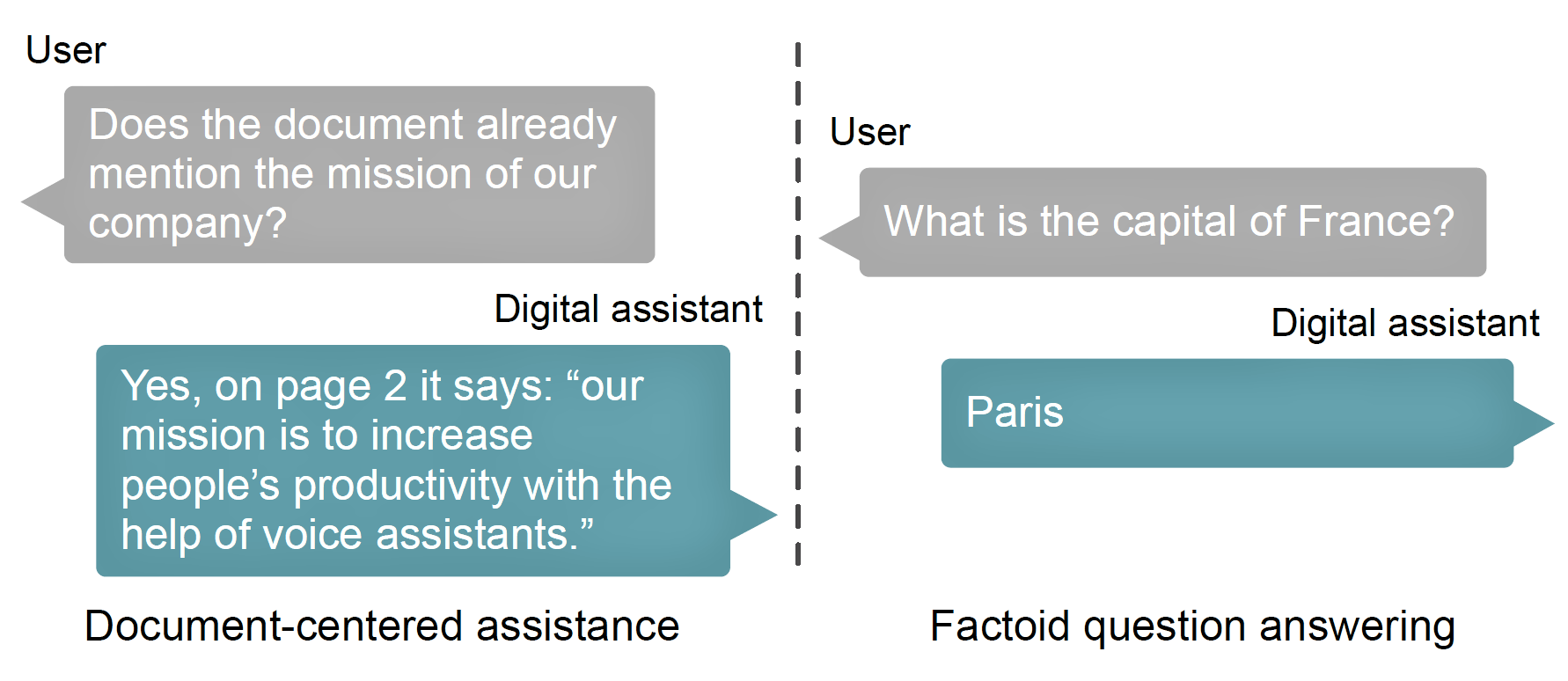}
	\caption{An example of document-centered assistance (left) vs. factoid question answering (right).}
	\label{fig:example_dialog}
\end{figure}

Digital assistants are used extensively to help people increase their productivity~\cite{voice_report}. A person can rely on their voice assistant, such as Amazon Alexa, Microsoft Cortana, or Google Assistant, to set an alarm while cooking, to play some music in the background, and to do a Web search on a recipe's ingredients. Conversational interaction is also playing an increasingly important role in helping people to increase their productivity for work-related tasks~\cite{ai_chatbots_intelligent_assistants_workplace}. 

One area of interest that has not seen significant progress is document-centered assistance. 
Consider the following example: a person is driving to a crucial business meeting to prepare for a day with potential investors. 
The person is co-authoring a document about their company that will be provided to its investors, and it will be finalized in the upcoming business meeting. 
To be optimally prepared for the meeting, the individual wants to review what is already in the document. 
Since they are driving, they do not have direct access to the document, so they call their conversational assistant. 
The assistant has access to the document and can answer any query related to the document. 
The driver might pose queries such as ``\textit{does the document mention the mission of our company?}'' or ``\textit{summarize what it says about our growth in the last two years.}'' -- queries that help them understand what is already outlined in the document and what they still have to add to finalize the document. 
At the same time, the driver is unlikely to ask factoid questions, such as ``\textit{who is the CEO of our company?},'' given that they are already familiar with the organisation. 
In fact, previous work in the context of email and Web search has shown that people's information needs are different when they are a co-owner of a document than when they are not~\cite{ai2017characterizing}. 
We hypothesize a similar difference in information needs in the context of document-assistance, motivated by the given example. 
This implies that document-centered assistance should critically differ from existing question answering (QA) systems, 
which are mostly trained to give short answers to factoid questions~\cite[e.g.,][]{rajpurkar2018know, reddy2019coqa}. 
Figure \ref{fig:example_dialog} gives an example of this difference. Document-centered assistance would also differ from non goal-oriented ``chit-chat'' scenarios~\cite[e.g.,][]{yan2018coupled, sankar2018modeling} -- in our document-centered scenario, people have very clear information needs. 

In this paper, we investigate this space of document-centered assistance. This is an important task, since good document-centered assistance has the potential to significantly increase a person's productivity. We specifically focus on text consumption and document comprehension scenarios in a work context, and we seek to answer the following three research questions:
\vspace{-\topsep}
\begin{enumerate}[leftmargin=*,label=(\textbf{RQ\arabic*})]
\item What kinds of conversational assistance would people like to receive in a document consumption scenario?
\item What kinds of queries might people use to receive this assistance when conversing with a document-aware assistant?
\item How well do initial baseline models do in a document-centered scenario?
\end{enumerate}
\vspace{-\topsep}
With this work we contribute:
\begin{enumerate}[leftmargin=*,label=(\textbf{C\arabic*}),nosep]
	\item An understanding of assistant capabilities that are important to enable the document consumption scenario;
	\item Insights into the types of questions people may ask in the context of document-centered assistance;
	\item A detailed exploration of a human-annotated dataset with: 
	\begin{enumerate*}
    	\item a collection of work-related documents,
        \item questions a person might ask about the documents, given some limited context,
        \item potential answers to the questions as represented by text spans in the document,
        \item additional metadata indicating some properties of the questions (for instance, it is a yes/no closed question, or the question is unanswerable given the document);
	\end{enumerate*}
	\item Baseline experiments applied to the dataset exploring ways to handle document-centered questions. 
\end{enumerate}
Our research consists of three steps:
\begin{enumerate*}
\item we perform a survey to answer RQ1 and RQ2 (Section~\ref{sec:user_study}); 
\item we proceed with a data collection step, outlined in Section~\ref{sec:data_collection}; and we answer RQ3 in Section~\ref{sec:modeling}.
\end{enumerate*}

%% file: 02-related_work.tex
%% !TEX root = ./main.tex

\section{Related Work}
\label{sec:related_work}
This paper is related to two broad strands of research. In the first part of this section we look into voice controlled document narration and natural language interactions with productivity software, which is relevant to the first step of our research, the survey. Our initial modeling steps focus on single-turn conversations, and so we conclude this section with work on question answering.

\subsection{Voice-Controlled Document Narration}
Document-centric assistance in the context of text-consumption is related to prior work that explores adding voice interactions to screen readers. Screen readers are accessibility tools that narrate the contents of screens and documents to people who are blind, or who have low-vision. In this space, \citet{ashok2015capti} implemented CaptiSpeak -- a voice-enabled screen reader that maps utterances to screen-reader commands and navigation modes (e.g., ``read the next heading'', ``click the submit button''). More recently, \citet{vtyurina2019verse} developed VERSE, a system that adds screen reader-like capabilities into a more contemporary virtual assistant. VERSE leverages a general knowledge-base to answer factoid questions (e.g., ``what is the capital of Washington''), but then differentiates itself by allowing users to navigate documents through voice (e.g., ``open the article and read the section headings''). An evaluation with 12 people who are blind found that VERSE meaningfully extended the capabilities of virtual assistants, but that the QA and document navigation capabilities were too disjoint -- participants expressed a strong interest in being able to ask questions about the retrieved documents. This strongly motivates the research presented in this paper.  

\subsection{Interactions with Productivity Software}
There is an increasing interest in how people use different devices for their work-related tasks \cite[e.g.,][]{karlson2010mobile, jokela2015diary, di2016surveying, williams2019mercury}. \citet{martelaro2019exploration} show that in-car assistants can help users to be more productive while commuting, yet in easy, non-distracting traffic scenarios. While digital assistance in cars is a recent development~\cite[e.g.,][]{lo2013development}, natural-language interfaces have existed for much longer in more traditional work scenarios; for example the search box in products such as Microsoft Office and Adobe Photoshop. \citet{bota2018characterizing} research search behavior in productivity software, specifically in Microsoft Office, and characterize the most used search commands. \citet{fourney2016automatic} investigate different types of queries users pose to a conversational assistant. Specifically they focus on \textit{semi implicit system queries} and \textit{fully implicit system queries}. They show that different types of queries can be reliably detected and that forms of query alteration can boost retrieval performance.

\subsection{Question Answering}
Question answering is the task of finding an answer to a question, given some context. 
A lot of progress has been made in the area, driven by the successful application of deep learning architectures and the increase of large scale datasets \cite[e.g.,][]{yang2015wikiqa, nguyen2016ms, rajpurkar2016squad, joshi2017triviaqa, trischler2016newsqa, dunn2017searchqa, rajpurkar2018know, kovcisky2018narrativeqa, yang2018hotpotqa, kwiatkowski2019natural}.  
Although these datasets are all unique, they mostly contain factoid questions that can be answered by short answer spans of only a few words. In addition, none of them contain queries that reference the document directly as the subject of the query, a distinction that can cause existing QA models to yield irrelevant or confusing responses. 

Considerable research has targeted neural QA \cite[e.g.,][]{bordes2015large, chen2017reading, dehghani2019learning, gan2019improving, kratzwald2019rankqa}. 
Recently, \citet{devlin2018bert} introduced BERT, or  Bidirectional Encoder Representations from Transformers. BERT is a language representation model that is pretrained to learn deep bidirectional representations from text. A pretrained BERT model can be fine-tuned on a specific task by adding an additional output layer. BERT has made a tremendous impact in many NLP tasks, including QA. In this paper, we base the baseline models on BERT transformers.

Some QA work has focused specifically on the low resource setting that we are also interested in in this work. Various approaches have been applied to augment small datasets to achieve good performance on language tasks (\cite[e.g.][]{lewis2019unsupervised,daniel2019towards,gan2019improving,yang2019data}). In order to accommodate our low-resource scenario, the data we have collected is supplemented with publicly available QA datasets.

\medskip\noindent%
All of the work cited above plays a role in setting context for our scenario. 
With the possible exception of VERSE, none have specifically explored how people might want to receive conversation-based assistance with documents, and in particular documents that they have rich context about. 
In the next section, we explore what features and queries users are most likely to pose to their assistant when a document is the focus of the conversation.

%% file: 03-user_study.tex
%% !TEX root = ./main.tex

\section{Step 1 -- Survey}
\label{sec:user_study}
In the first step of our research, we aim to answer (\textbf{RQ1}) \textit{What kinds of conversational assistance would people like to receive in a document consumption scenario?}, and (\textbf{RQ2}) \textit{What kinds of queries might people use to receive this assistance when conversing with a document-aware assistant?} To do so, we conduct a survey to explore the space of queries that people might pose when communicating with a voice assistant about a document, while not having full access to this document. We focus on a consumption scenario while on the go (i.e.,\ limited primarily to voice and some touch input/output). Specifically, participants in our survey are presented with the following scenario: ``\textit{You are on your way to a business meeting. To help you prepare, your manager has sent you an email with a document attached. The objective of the meeting is to finalize this document, so that it can be shared with the rest of the organization. Your manager's email also includes the introduction of the document. You have been able to read this introduction, so you have an idea what to expect. You have not read the full document yet, but you can assume the document is approximately 6 pages long. On your way to the business meeting you do not have time to access the document, but you do have your smartphone equipped with a voice assistant like Alexa, Google Assistant, or Cortana. The voice assistant can help you navigate and understand what is written in the document, so that you will arrive prepared at your meeting. The voice assistant can answer your questions via audio or by displaying information on your smartphone screen.}''

\subsection{Survey Overview}
\begin{figure}
	\centering
	\includegraphics[width=0.85\columnwidth]{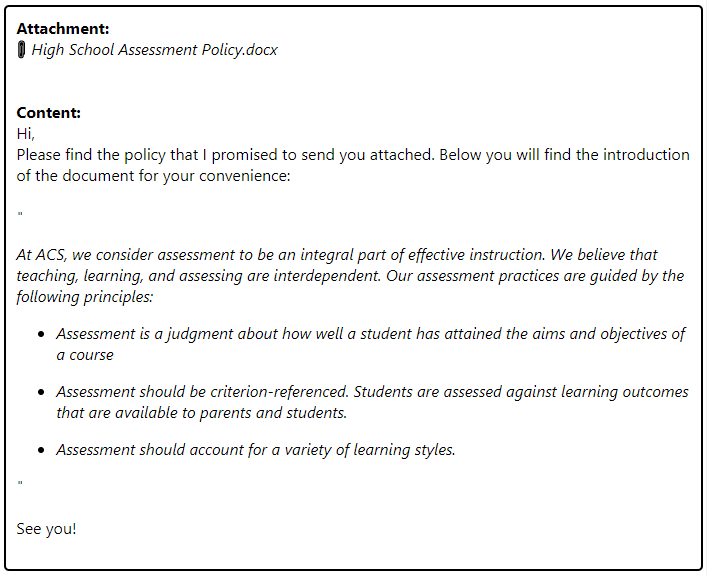}
	\caption{Sample e-mail used to inform participants.}
	\label{fig:user_study_email}
\end{figure}

Our survey consisted of two parts, corresponding to RQ1 and RQ2. In the first part, our primary goal was to explore three sub questions: (1) do users recognize the outlined scenario as relevant to their daily lives?, (2) would users find voice assistance in the outlined scenario helpful?, and (3) what range of features are important to users in a voice-first document consumption scenario? Having identified the range of functionalities that a document-centered conversation might cover, in part two of our survey we aimed to gain a better understanding of the types of questions users might ask.
Therefore, we collected questions that are grounded in specific documents. To this end, participants were primed with the same scenario as in the first part. The scenario is simulated by presenting them with an email that mimicked the email they received from their manager while on the go. The email contained the document introduction as a means to give them context about a specific document, to ensure that participants were able to ask informed questions, yet did not have full knowledge about what is written in the document. Figure \ref{fig:user_study_email} shows an example of an email provided to participants.

\subsection{Participants}
Our task was performed by $23$ participants in a judging environment comparable to Amazon Mechanical Turk.\footnote{https://www.mturk.com/} Participants were all English speaking and U.S.-based. Participants were paid at an hourly rate, removing the incentive to rush responses. We did set a maximum time of ten minutes per document.

\emph{Instructions given to participants.}
Before the task, participants were provided with detailed guidelines of the task and trained to follow them. In these guidelines, we explicitly encouraged participants to ask questions that were document-centered, i.e., to closely keep the outlined scenario in mind when asking questions. The participants were instructed to avoid questions that might be posed about any document, and answered using more mechanical solutions (e.g., \textit{who is the author?}, \textit{how many pages?}), and steered towards a scenario where they imagined having some familiarity with the document subject.
Although we acknowledge that these more general questions are highly relevant, we argue that we do not need many sample questions of this type to fully understand the space of potentially relevant mechanical questions. Note that in the first part of the survey we investigated what participants would find the most and least important features in the outlined scenario, and this gives them the opportunity to select more mechanical features.
Participants were explicitly told to imagine their ideal voice assistant and to not limit themselves by any prior assumptions about the capabilities of currently existing voice assistants.

\emph{Participant training.}
Participants performed two training rounds, after which we provided them with feedback on their constructed questions in part two. This way we aimed to ensure that participants understood the task and devise high quality responses.

\subsection{Document Selection}
\label{subsec:document_selection}
We selected $20$ documents from a larger data set of $615$ documents in Microsoft Word format. These documents were retrieved from a broad crawl of the Web and meet the requirements that they are written in English and can be easily summarized. This last requirement, which was manually verified, ensures that we have a high quality dataset where noisy documents such as online forms are excluded.
We selected the $20$ documents from this set based on:
\begin{enumerate*}
\item the document should contain a clear introduction;
\item the document should be between $3$ and $10$ pages long; and
\item the topic of the document should be understandable for non-experts on this topic and should not be offensive to anyone. \end{enumerate*}
Table \ref{tab:user_study_doc_overview} gives more details on the nature of the selected documents. In addition to these $20$ documents, we chose another two documents with which to train the participants. Although slightly deviating from the co-ownership scenario, providing users with documents ourselves allowed us to collect data in a more structured way, which we can use for the remaining research questions at a later stage. In the second part of the survey, the question collection round, each participant was asked to pose five questions about a given document. We required $20$ judges per document. Since we have $20$ documents we acquire $400$ human intelligence tasks (``HITs''), resulting in $2000$ questions.

\begin{table}
	\caption{Categories of selected documents (20 in total) and their frequency in the survey distributed to participants.}
	\label{tab:user_study_doc_overview}
	\begin{tabularx}{0.8\columnwidth}{lr}
	\toprule
		\textbf{Document category} & \textbf{Document count} \\ \midrule
		Report & 3 \\
		Job application & 3 \\
		Description of a service & 3\\
		General description & 2\\
		Guidelines & 3\\
		Policy & 3\\
		Informative / Factsheet & 3\\ \bottomrule
	\end{tabularx}
\end{table}

\subsection{Survey Results}
In this section, we provide the precise formulation of our survey questions, as well as the participants' responses to these questions.

\krimp
\subsubsection{Part 1 -- Survey Questions}

\begin{figure}
\centering
\begin{subfigure}{0.5\textwidth}
  \centering
  \includegraphics[clip,trim=3mm 5mm 0mm 3mm,width=0.9\textwidth]{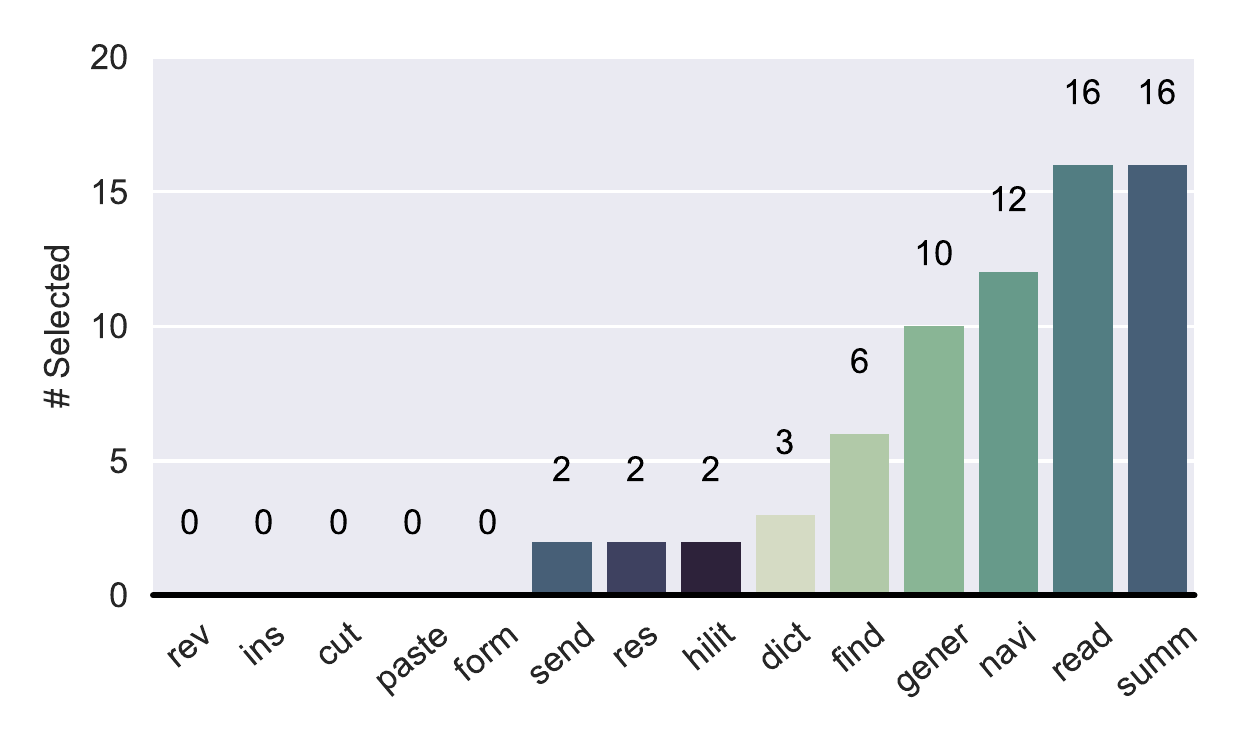}
  \vspace{-2mm}
  \caption{Responses to question 3 -- most useful assistant capabilities.}
  \label{fig:most_useful_features}
\end{subfigure}
\begin{subfigure}{0.5\textwidth}
  \centering
  \includegraphics[clip,trim=3mm 5mm 0mm 3mm,width=0.9\textwidth]{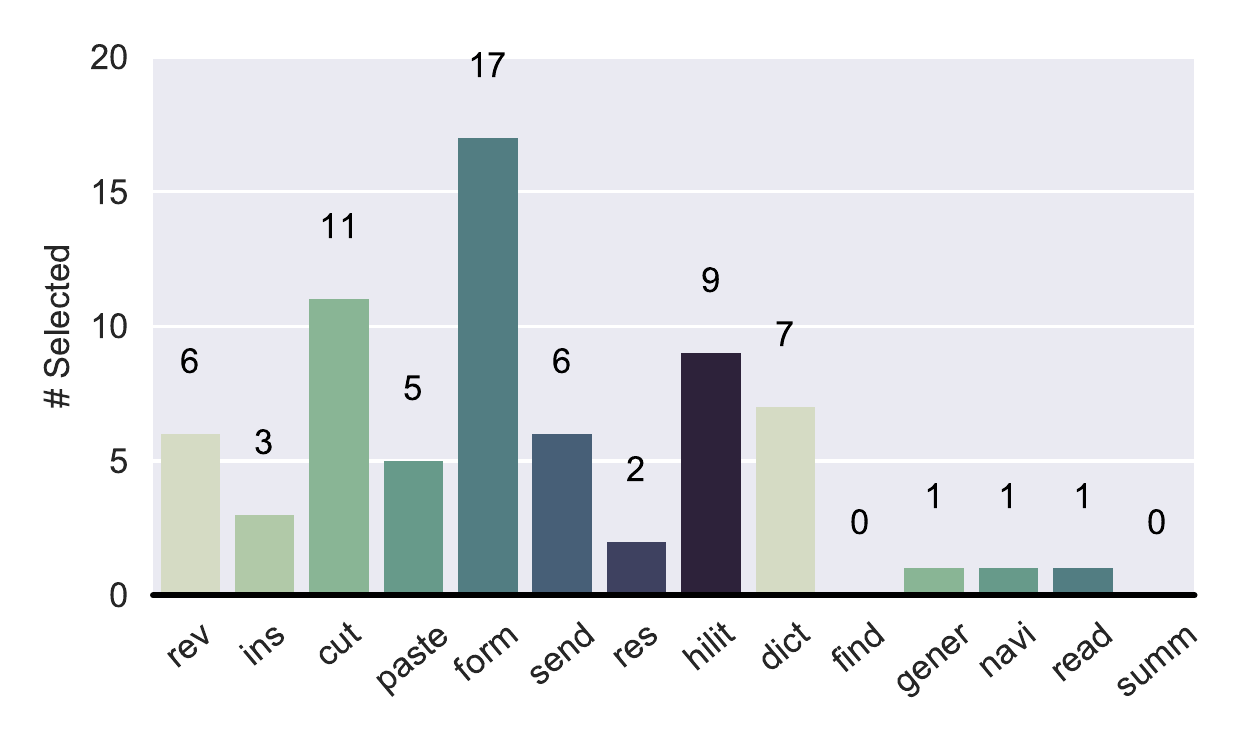}
  \vspace{-2mm}
  \caption{Responses to question 4 -- least useful assistant capabilities.}
  \label{fig:least_useful_features}
\end{subfigure}
\vspace*{1mm}
\caption{Most and least useful assistant capabilities; names explained in Table~\ref{tab:feature_options}. On the y-axis: the number of times this particular capability was selected by participants (max = 23).}
\label{fig:feature_usefulness}
\end{figure}

\begin{enumerate}[wide, labelwidth=!, labelindent=0pt,nosep]
    \item \textit{Do you recognize the outlined scenario (i.e., needing to quickly catch up on a document while on the go) or some variation of it as something you experience in your daily life?}

	\noindent $22$ out of $23$ participants indicated that they recognized the scenario.

	\item \textit{Do you expect to find it helpful if a voice assistant helps you to quickly familiarize yourself with the document in the outlined scenario?}

	\noindent $22$ out of $23$ participants indicated that they would find this helpful.

	\item \textit{From the list below, choose three capabilities that you would find \textbf{most useful} in a voice-powered AI assistant to help prepare you for the meeting.}

	\noindent Participants could choose from the capabilities listed in Table \ref{tab:feature_options}. We randomized the order in which the features were presented, to avoid position biases. Note that the prompt specifically references the consumption scenario that participants are primed to consider. The results are given in Figure~\ref{fig:most_useful_features}. Please refer to Table~\ref{tab:feature_options} to match the abbreviation on the x-axis with the feature description.

	\item \textit{From the list below, choose three capabilities that you would find \textbf{least useful} in a voice-powered AI assistant to help prepare you for the meeting.}

	\noindent Again, participants could choose from the capabilities in Table \ref{tab:feature_options} and again this list is randomized for each participant. Figure \ref{fig:least_useful_features} shows the results for this question. Comparing the results in Figure~\ref{fig:most_useful_features} and Figure~\ref{fig:least_useful_features} shows that participants are very consistent in the capabilities they find most and least useful.

	\item \textit{Can you think of any other features that you would like the voice assistant to be capable of? Please describe.}

	\noindent We divided the participants' answers into ``mechanical'' features and ``overview'' features. A sample of the answers is presented below.

	\noindent \textit{Mechanical features:}
	\begin{itemize}[leftmargin=*]
	    \item ``Voice recognition to unlock phone''
        \item ``Automatic spelling and grammar check''
        \item ``Remind me where I stopped when reading''
        \item ``The ability to link another app, such as maps or notes to the document directly''
        \item ``Bookmarking specific sections for future reference''
        \item ``Another useful feature would be the ability to add highlighted text to multiple programs simultaneously such as email notes and any other app''
        \item ``The Assistant should be able to turn tracked changes on and off and accept/reject changes and clean up a document and finalize''
    \end{itemize}

    \noindent \textit{Overview features:}
    \begin{itemize}[leftmargin=*]
        \item ``Give bullet points of main topics''
        \item ``Give information about key points''
        \item ``Just highlight key points, summarize document''
        \item ``I would like for the voice assistant to be able to pick out the main points and read them out to me via voice output''
        \item ``If the assistant was able to give a synopsis then ask 1 or 2 questions to be sure the user understands the info''
    \end{itemize}
\end{enumerate}

\begin{table}
\caption{Assistant capabilities suggested to participants and judged for their utility. Abbreviations were never shown to users and are only used to map plots in this paper to the corresponding capability.}
\label{tab:feature_options}
\begin{tabularx}{\columnwidth}{lX}
\toprule
	\textbf{Abbr.} & \textbf{Capability} \\ \midrule
	cut & Cut content from the document using voice \\
	dict & Dictate input to the document \\
	find & Find specific text in the document using voice input \\
	form & Change text formatting using voice \\
	gener & Respond to general questions about the document content, using voice input and output \\
	hilit & Highlight text using voice \\
	ins & Insert new comments into the document using voice \\
	navi & Navigate to a specific section in the document using voice input \\
	paste & Paste content from the device clipboard using voice \\
	read & Read out the document, or parts of it, using voice output \\
	res & Respond to existing comments in the document using voice \\
	rev & Revise a section of text using voice input \\
	send & Send or share a section of text using voice input \\
	sum & Summarize the document, or parts of it, using voice output \\
	\bottomrule
\end{tabularx}
\end{table}

\krimp
\subsubsection{Part 2 -- Collecting Questions}
\begin{figure}
	\centering
	\includegraphics[width=\columnwidth]{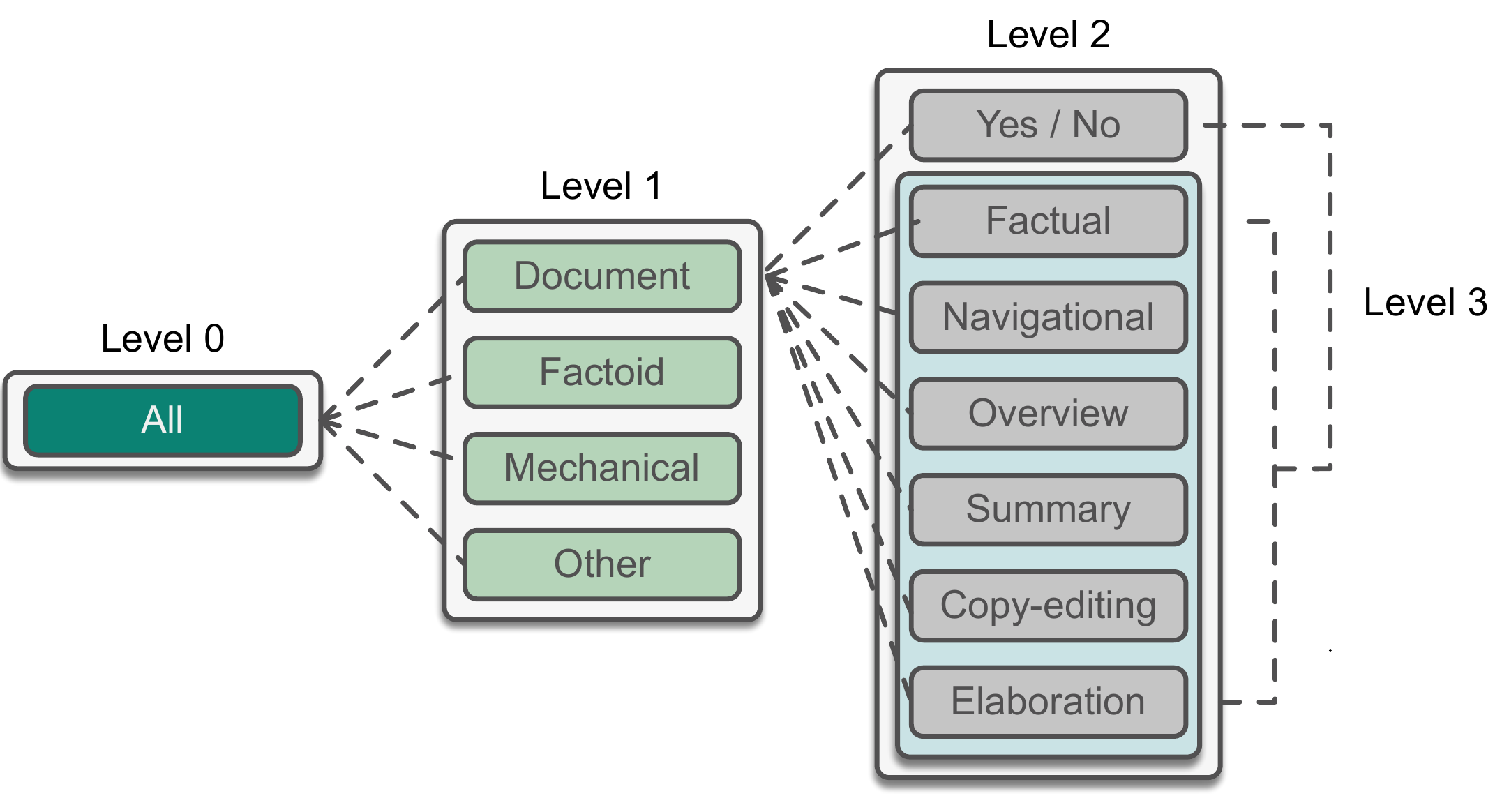}
	\caption{Question hierarchy.}
	\label{fig:question_hierachy}
\end{figure}

\begin{figure*}
\centering
\begin{subfigure}{\textwidth}
  \centering
  \includegraphics[clip,trim=0mm 7mm 0mm 0mm 0mm,width=\linewidth]{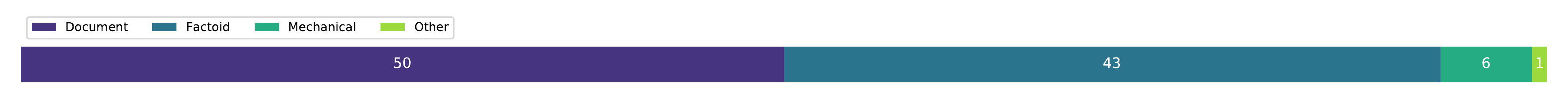}

  \vspace*{-2mm}
  \caption{Distribution level 1 question types (\%).}
  \label{fig:level_1_question_types}
\end{subfigure}
\begin{subfigure}{\textwidth}
  \centering
  \includegraphics[clip,trim=0mm 7mm 0mm 0mm 0mm,width=\linewidth]{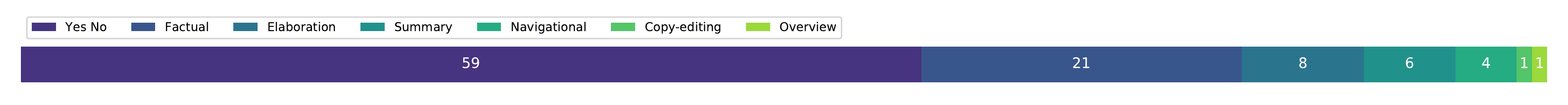}

  \vspace*{-2mm}
  \caption{Distribution level 2 question types (\%).}
  \label{fig:level_2_question_types}
\end{subfigure}
\begin{subfigure}{\textwidth}
  \centering
  \includegraphics[clip,trim=0mm 7mm 0mm 0mm 0mm,width=\linewidth]{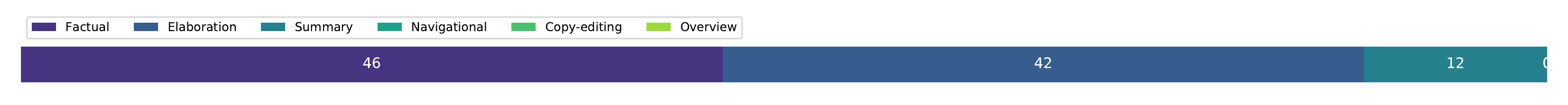}

  \vspace*{-2mm}
  \caption{Distribution level 3 question types (\%).}
  \label{fig:level_3_question_types}
\end{subfigure}
\vspace*{0mm}
\caption{Distribution of question types per hierarchical level. (Best viewed in color.)}
\label{fig:question_types}
\vspace{-10pt}
\end{figure*}

Here we present the results of the second part of the survey, in which participants were prompted to generate questions about a document.
Recall that the participants were only shown the document introduction or preamble and did not have visibility into the full document text.
\begin{enumerate}[wide, labelwidth=!, labelindent=0pt]
    \setcounter{enumi}{5}
    \item \textit{Please ask five questions to your voice assistant that would help you understand what is written in the document.}

    \noindent We can divide participants' answers into a hierarchy of question categories. Note that the responses can be both questions and directives (e.g., ``\textit{go to Section X}''). Since the vast majority of the collected responses are questions, for brevity we refer to both of these response types as \emph{questions}. Figure \ref{fig:question_hierachy} shows the hierarchy. It was developed by sampling a set of participants' questions, which an expert studied and categorized. Three experts then reviewed all questions and categorized them according to the proposed taxonomy. By reviewing where the experts disagreed, some minor adjustments were made to the hierarchy to arrive at the final one shown here.
    Level 1 of the hierarchy corresponds to how the question can be best responded to, or what kind of system or model would be suited best to handle the questions. Because document-centered questions are the main interest of our current research, we divide those into another set of categories, describing the intents of users on this level in more detail. This is level 2. We also subdivide the yes / no questions into the rest of the categories of level 2 and call this level 3. We do this because it is questionable whether a person would really be satisfied with a simple ``\textit{yes}'' or ``\textit{no}'' in response. We describe the question types in Table~\ref{tab:question_hierarchy_examples}, and also provide verbatim examples sourced from the participants' responses. Figure~\ref{fig:level_1_question_types} shows the distribution of question categorizations on level~1. Document-centered questions form the largest category of the questions. Recall that participants had to ask $5$ questions per document; we investigated whether these questions differed in type. E.g., did participants ask mechanical questions first (``\textit{bring me to Section 2.}'') and then a document-centered question (``\textit{what does it say there about X?}'')? We did not find such a difference. We also investigated whether the type of document (Table \ref{tab:user_study_doc_overview}) was an indication for the types of questions that were asked, but found no difference between document types. The user was a strong indication for the type of question that was asked, indicating varying interpretations of the outlined scenario. Some users ask only factoid questions, some users only ask document-centered questions and only a few ask a mixture of all question types.

    The division of category labels for level~2 is shown in Figure \ref{fig:level_2_question_types}. As can be seen, the majority of questions are closed form yes / no questions. Figure \ref{fig:level_3_question_types} shows how these questions were categorized on level~3, yielding only $3$ copy-editing questions, $2$ overview questions, and $1$ navigational question, rounding down to $0\%$ in Figure \ref{fig:level_3_question_types}.
\end{enumerate}

\begin{table*}
    \centering
    \caption{Question type descriptions and examples.}
    \label{tab:question_hierarchy_examples}
    \begin{tabular}{lp{7.8cm}p{7.8cm}}
    \toprule
        \bf Level & \bf Question type & \bf Examples \\
        \midrule
        \multirow{12}{*}{L1} & \textbf{Document:} These are document-centered questions. That is, the question's phrasing explicitly or implicitly references the document. When asking such a question, a user is not looking for encyclopedic knowledge, yet rather for assistance that can help them to author the document. These types of questions are not present in existing QA datasets. & Does the document have specifications to the type of activity and sector improvement that will be offered? \\
        \cmidrule{2-3}
        & \textbf{Factoid:} Fact-oriented question that co-owners of a document are unlikely to ask. Answers are often only a few words long. Existing QA datasets cover these types of questions very well. & What is the date of the festival? \\
        \cmidrule{2-3}
        & \textbf{Mechanical:} Questions that can be answered with simple rule-based systems. & Highlight ``Capability workers'' \\
        \cmidrule{2-3}
        & \textbf{Other:} Questions that fall outside the above categories. & Read the email to me.
        \\
        \midrule
        \multirow{13}{*}{L2} & \textbf{Yes / No:} Closed form (can be answered with `yes' or `no'). & Does the document state who is teaching the course? \\
        \cmidrule{2-3}
        & \textbf{Factual:} Questions that can be answered by returning a short statement or span extracted from the document. & Where does the document state study was done? \\
        \cmidrule{2-3}
        & \textbf{Navigational:} Referring to position(s) in the document. & Go to policies and priorities in the doc. \\
        \cmidrule{2-3}
        & \textbf{Overview:} Questions that refer to the aim of the document.& What is the overall focus of the article? \\
        \cmidrule{2-3}
        & \textbf{Summary:} Questions that ask for a summary of the document or of a particular part of the document.& Find and summarize coaching principles in the document. \\
        \cmidrule{2-3}
        &\textbf{Copy-editing:} Questions when editing a document. They require a good understanding of the document to answer. & Highlight text related to application of epidemiologic principles in the document \\
        \cmidrule{2-3}
        & \textbf{Elaboration:} Questions that require complex reasoning and often involve a longer response.& Please detail the process to get access to grant funds prior to confirmation. \\
    \bottomrule
    \end{tabular}
\vspace{-5pt}
\end{table*}

\subsection{Classifying Question Types}
We trained a simple, yet effective logistic regression classifier to classify the question types. From Table \ref{tab:classification_results} it becomes clear that we can accurately learn to classify different question types, especially at higher levels in the hierarchy. These labels are extremely helpful for a number of tasks: they are useful to decide what type of answer the user is expecting, or the type of model that should deliver a response. An accurate classification on the first level is important for this task: do we want to use a rule-based system, a factoid QA model, or a newly trained document-centered QA model? The results on the second level can be used to decide whether or not we face a yes / no question and therefore may have to start the answer with ``yes'' or ``no.'' In a question generation setting, the labels can also be used to condition the question generation process.

\begin{table}
\caption{Question type classification results. Mean accuracy and variance after 5-fold cross validation.}
\label{tab:classification_results}
\begin{tabular}{r@{~}l r@{~}l r@{~}l}
\toprule
	\multicolumn{2}{c}{\textbf{Level 1}} & \multicolumn{2}{c}{\textbf{Level 2}} & \multicolumn{2}{c}{\textbf{Level 3}} \\
	\midrule
	$0.92$& $(\pm 8.6e^{-5})$ & $0.90$ & $(\pm 1.3e^{-4})$ & $0.67$ & $(\pm 1.0e^{-3})$ \\
	\bottomrule
\end{tabular}
\end{table}

\subsection{Answering RQ1 and RQ2}
The results of the survey allow us to answer our first two research questions. We have identified a range of capabilities that users would like to see in a document-centered assistance scenario, and we have identified a hierarchy of questions that users would ask. Document-centered questions are different from factoid QA questions and form an interesting new category of questions to research.

%% file: 04-QA_data_collection.tex
%% !TEX root = ./main.tex

\section{Step 2 -- Data Collection}
\label{sec:data_collection}
The first step of our work shows that users pose different types of questions to a digital assistant when seeking document-centered assistance than are typically present in modern QA datasets. To dive deeper, we first scale up our data collection to gather more questions and proposed answers to those questions. In this section, we describe our data collection process and the statistics of the collected data. We refer to the collected data as ``DQA'' dataset, short for Document Question Answering.

\subsection{Question Collection}
For question collection, we randomly selected another $36$ documents (recall Section \ref{subsec:document_selection}) using the same selection criteria. We asked the same set of participants as in Step 1, now acting as crowd workers, to generate questions for these documents. This time we omitted the survey questions about the scenario and capabilities; we asked them to pose five questions about the document. Since we only presented the workers with the document introduction, it is likely that workers will also ask questions that cannot be answered from the document, more closely resembling a real life situation.

\subsection{Answer Collection}
Once we collected the questions, we asked the same pool of crowd workers to select answers for these questions. We presented workers with the full document and asked them to read it carefully. Then we asked them to answer five questions about the document. These questions were always a set of five questions that were asked by one of the crowd workers in the question collection round (not necessarily the same as the worker who is answering the questions). The questions were kept together and were presented in the same order as they were asked, due to the potential conversational nature of the questions. Note that this is only applicable to a few instances in the data, allowing us to train a single-turn QA model later. 
Each set of questions is answered by three crowd workers. Figure \ref{fig:sample_QA_overview} in Appendix \ref{sec:appendix_A} shows an overview of the presented task.

\begin{figure*}
	\centering
	\includegraphics[width=0.75\textwidth]{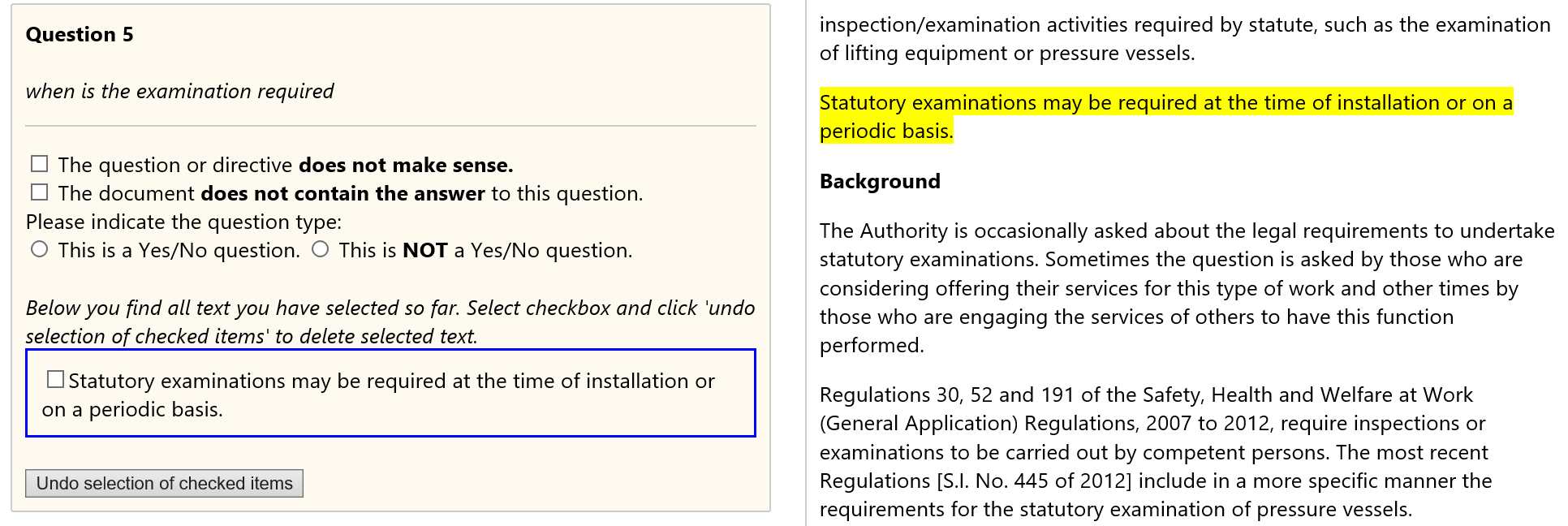}
	\caption{Question answering data collection selected text. (Best viewed in color.)}
	\label{fig:sample_QA_selection}
	\vspace{-10pt}
\end{figure*}

For each question, we display the following options after a click on the question:
\begin{enumerate*}
    \item This question or directive does not make sense; 
    \item The document does not contain the answer to this question; and
    \item Please indicate the question type: (a) This is a yes / no question, (b) This is not a yes / no question.
\end{enumerate*} 
If a worker selects that the question is a yes / no question, we ask them to indicate whether the answer is ``yes'' or ``no'' and to select parts of the document with supporting evidence. If no supporting evidence could be found in the document (e.g., because the question was \textit{``does the document contain information about topic X?''} and the answer was ``no'') we asked workers to tick the box that supporting evidence cannot be highlighted. Figure \ref{fig:sample_QA_yesno} in Appendix \ref{sec:appendix_A} shows an example of the task including the expansion that is shown if a worker selects that the question is a yes / no question. If the worker has not clicked any of the above mentioned options, it means the question is valid, open-ended, and answerable. For these questions, we asked workers to select the minimal spans of text necessary to answer the question. Workers could select up to three spans in the document; each span was at most $700$ characters in length. Since some documents can be challenging to understand, we included a checkbox to indicate that the questions were difficult to answer or the document was hard to understand. Figure \ref{fig:sample_QA_selection} shows an example of the highlighting tool. Text highlighted in the document (right-hand pane), is populated as a selected span in the left-hand pane (blue box).

We again performed $2$ training rounds with the crowd workers, in which we ensured workers fully understand the task. During the data collection phase an expert spot-checked answer quality.

\subsection{Dataset Statistics}
Table ~\ref{tab:answer_question_types} describes the distribution of annotations about the questions that were collected from the crowd workers. Recall that each question was judged and answered by $3$ workers. Here we present the raw numbers.

During the question generation phase workers were not shown the full document, whereas the workers have access to the full text while selecting answers. This disparity is reflected in the statistic that 40\% of questions were considered unanswerable from the text. This ensures that our dataset is suitable for training a system that can identify unanswerable questions.

\begin{table}
\caption{Answer and question types.}
\label{tab:answer_question_types}
\begin{tabular}{lrr}
\toprule
	\textbf{} & \textbf{Number} & \textbf{\% (of total)} \\ \midrule
	Annotated documents & 56 & -- \\
	Valid questions (= annotation tasks) &  16375 & 100.00\\ 
	Invalid questions (discarded) & 425 & -- \\
	Open questions &  9442 & 57.66 \\
	Yes/no questions & 6933 & 42.34 \\
    No answer & 6543 & 39.96 \\
    No evidence & 1748 & 25.21 \\
	\bottomrule	
\end{tabular}
\end{table}

Table ~\ref{tab:span_statistics} gives an overview of the number of spans and the lengths of spans that were selected by crowd workers. The average span length is substantially larger than the average span length of only a few words in most existing QA datasets. This supports our claim that the current document-centered scenario requires different types of data to train on. Table~\ref{tab:agreement_statistics} describes the distribution of annotation responses, in particular the fraction of questions where workers were in full agreement about the impossibility of answering a question from the text (52\%) (random full agreement would be 25\%), as well as ROUGE-\*scores describing the mean self-similarity of selected spans across judges who responded to the same question. 
Hence, participants agreed well with each other.

\begin{table}
\caption{Span statistics. Span length in tokens.}
\label{tab:span_statistics}
\begin{tabular}{l@{}r}
\toprule
	\textbf{} & \textbf{Statistic} \\ \midrule
	Total spans &  11702\phantom{.000} \\
	Average number of spans per question (all) & 0.715 \\
	Average number of spans per question with answer &  1.45\phantom{0} \\
	Average span length per question (all) & 26.69\phantom{0} \\
	Average span length per question with answer & 37.35\phantom{0} \\
	\bottomrule	
\end{tabular}
\end{table}

\begin{table}
\caption{Agreement statistics.}
\label{tab:agreement_statistics}
\begin{tabular}{ll@{~~}l}%{lr@{~~}l}
\toprule
	\textbf{} & \textbf{Metric} & \\ \midrule
	Impossible full agreement (\%) & 52.09 & \\
	Impossible partial agreement (\%) &  47.91 & \\
	Rouge-1 F-score avg (questions with span) & 52.44 & ($\pm 8.79$) \\
	Rouge-2 F-score avg (questions with span) & 44.92 & ($\pm 11.14$) \\
	Rouge-L F-score avg (questions with span) & 46.89 & ($\pm 9.54$)\\
	\bottomrule	
\end{tabular}
\end{table}

%% file: 05-modelling.tex
%% !TEX root = ./main.tex

\section{Step 3 -- Baseline modeling}
\label{sec:modeling}

We present baseline models for passage retrieval and answer selection on our dataset. Our aim is to answer (\textbf{RQ3}) \textit{How well do initial baseline models do in a document-centered scenario?}

\subsection{Data Preprocessing}
We use exactly the same format as the popular SQuAD2.0 \cite{rajpurkar2018know} dataset for our preprocessing output. We keep all questions and answers for a random sample of $25\%$ of the documents as a separate hold-out set. 
Recall that we have collected $3$ answers per question, as we had $3$ workers answer each question. We discarded all invalid questions and we ensured that the remaining labels (such as ``yes / no questions'') were consistent as follows. First we looked at workers' answers for whether the question was a yes / no question and computed the majority vote. We kept the answers of the workers who agreed with the majority vote and discarded the rest (if any). The majority vote has been shown to be a strong indication for the true label~\cite{li2019probabilistic}. In case of a tie, we chose to treat this question as a yes / no question as it provided us with most information about the question, which is beneficial for training. If the question is now labeled as a yes / no question we continue to the answer (i.e., ``yes'' or ``no''). Again we computed the majority vote and only kept the answers from workers who agreed with the majority vote. In case of a tie we chose ``yes'' as the answer, as this results in the richest label for the question. Then we followed the same procedure for the ``no-evidence'' checkbox, choosing to include spans in the event of a tie. 
Lastly, if the question was not labeled as a yes / no question, we applied the same majority vote and tie-breaking strategy for whether the document contains the answer. 
Using this approach, we kept approximately half of the collected question-answer pairs, but ensured that no model is trained on contradictory answers. This improved model performance. During training, we used the collected question-answer pairs as individual training examples, i.e.,\ if we have $2$ answers for a question given by $2$ workers, we added them separately to our training set. This way we increased the number of training samples. At this stage, we also chose to add all selected spans for an answer separately to our training set. We leave multiple span selection for future work. During evaluation we treated all selected answers for a question as valid answers.

\subsection{Passage Ranking}

In this section, we describe our approach for initial passage ranking experiments on our new DQA dataset. We explore three baseline methods: random selection, BM25-based ranking, and selecting the first passage in the document.

\krimp
\subsubsection{Passage Construction}
During data collection, crowd workers selected answers to questions, yet they did not select the paragraphs or passages that include these answers. Therefore we constructed passages for all questions with answers as follows. We discard questions without answers in this experiment. We split each document in the dataset into sentences. We adopted a sliding window approach, moving our window one sentence at the time, constructing passages of size \textit{window size}. We set the window size to $5$. We also divided the selected answers into sentence chunks (or smaller, if only parts of sentences were selected). For each answer, we scored each passage by the number of chunks it contains. That is, a passage received a point for each chunk that is also in the answer.

\krimp
\subsubsection{Baseline Passage Ranking 1 -- Random}
For this baseline we retrieve a random passage. For each retrieved passage we compute the ROUGE-1 F-score, ROUGE-2 F-score, and ROUGE-L F-score (based on retrieved passage and ground truth)~\cite{lin2004rouge} and the \textit{Precision@1}.
Recall that we scored paragraphs based on the number of overlapping chunks with the selected answer. Therefore some paragraphs contain only part of the answer, and some contain the full answer. To account for this difference we computed a so-called \textit{hard} and \textit{soft} \textit{Precision@1}. For the hard version, we assigned binary labels to retrieved passages; $1$ if the retrieved passage contains (part of) the answer, $0$ if it does not. For the soft version, we scored each retrieved passage as follows: we took the number of overlapping chunks of the retrieved passage and the answer and divided this by the maximum number of overlapping chunks. Since annotators may select answers from different passages, we optimistically took the best passage score per question, i.e., we returned a valid match if the selected passage matched any annotator response.

\krimp
\subsubsection{Baseline Passage Ranking 2 -- First passage}
For this baseline, we select the document's first passage as an answer to each question. We compute the same metrics as in Baseline 1. The purpose of this baseline is to establish to what extent answers to questions are biased by their presence in the preamble of the document, which was shown to study participants at question generation time.

\krimp
\subsubsection{Baseline Passage Ranking 3 -- BM25}
For this baseline, we retrieve the best matching passage with BM25~\cite{robertson2009probabilistic} and compute the same metrics as in Baselines $1$ and $2$.

\krimp
\subsection{Results for Passage Ranking}
In Table \ref{tab:results_par_ranking} the results for the passage ranking experiments are shown. Analysis of variance (ANOVA), $F(2,54)>8.9$, $p<0.0002$ yields significant differences between the three approaches. A post-hoc Tukey test $p<0.05$ shows that first passage selection significantly outperforms Random for all measures, and BM25 for all measures except ROUGE-L. BM25 significantly outperforms Random only for ROUGE-L. We hypothesize that the performance of first passage selection can have a number of causes: (1) because workers have been shown the introduction of the document, many questions can be tailored towards information located in the introduction, (2) workers have read the document from beginning to end, which may have biased them towards selecting from the first part of the document and not from the later parts once they found the answer.

\begin{table}
\caption{Results for passage ranking.}
\label{tab:results_par_ranking}
\begin{tabular}{lccccc}
\toprule
	\textbf{Model} & \textbf{P@1} & \textbf{P@1} & \textbf{Rouge-1} &  \textbf{Rouge-2} & \textbf{Rouge-L} \\
	& \textbf{soft} & \textbf{hard} & \textbf{F-score} & \textbf{F-score}  & \textbf{F-score} \\ \midrule
	Random & 0.29 & 0.31 & 0.26 & 0.13 & 0.19 \\
	First & 0.55 & 0.56 & 0.32 & 0.20 & 0.23 \\
	BM25 &  0.32 & 0.34 & 0.29 & 0.16 & 0.22 \\

	\bottomrule	
\end{tabular}
\end{table}

\subsection{Answer Selection}
In this section, we discuss how state-of-the-art models for answer selection perform on the DQA data and DQA enhanced with data from the SQuAD2.0 dataset~\cite{rajpurkar2018know}. We select this dataset for two reasons: first, it is a standard dataset for benchmarking Question Answering tasks and, second, like DQA, it contains questions marked as unanswerable, making it closely compatible with our collected data. All baselines were evaluated using the DQA hold-out set. 

\krimp
\subsubsection{Passage Construction}
For the answer selection experiments, we selected the passages for each answer using the same windowing method as in the passage ranking experiments. The only difference is that we now only considered passages that contain the full answer. For unanswerable questions, we selected the best matching paragraph with BM25. Even though our previous experiments showed that the answer is often in the first paragraph, we chose BM25 as a less biased and more informed selection procedure.

\krimp
\subsubsection{Baseline Answer Selection 1 -- Fine-tuned BERT on SQuAD2.0}
For QA,  BERT is fine-tuned as follows. A question and a passage are fed to a pre-trained BERT language model. They are separated with a separator token. The final output layer is trained to select the start and end index of the answer, from the input passage. If no answer is detected in the passage, $0$ is selected as index for both start and end. For the current baseline we fine-tuned HuggingFace's  implementation of BERT Large~\cite{Wolf2019HuggingFacesTS} 
on 8 Titan XP GPUs, using SQuAD$2.0$. First, we ensured we got similar scores as reported in the repository for the SQuAD2.0 tasks. Then, we evaluated the model on the DQA hold-out set. We included this baseline to test how a pretrained and fine-tuned BERT model on a very popular QA dataset performed on our DQA dataset without any adaption.

\krimp
\subsubsection{Baseline Answer Selection 2 -- Fine-tuning on SQuAD2.0 with query rewriting}
For this baseline, we used the same fine-tuned BERT model as for Baseline 1, yet this time we performed some simple query rewriting on the hold-out set to make our questions more comparable to those the model is fine-tuned on. For query rewriting, we computed the most common n-grams in our document train set. We manually inspected those n-grams and chose to delete the following document and conversational related patterns from our questions, expressed as Python regular expressions:

\begin{itemize}[leftmargin=*,nosep]
    \item \verb|'^does( the)? document (\S)+ (you)? '|
    \item \verb|'^does it (\S)+ '|
    \item \verb|'^what does( the)? document (\S)+ (you)? ')|
    \item \verb$'according to( the)? document(\s,\s|,\s|\s)')$  
    \item \verb|'in( the)? document '|
    \item \verb|'^assistant, '|
\end{itemize}

\krimp
\subsubsection{Baseline Answer Selection 3 -- Fine-tuning on DQA}
For this experiment, we fine-tuned BERT Large using the DQA dataset, again used the same fine-tuning implementation as used previously. We evaluated on the DQA hold-out dataset.

\krimp
\subsubsection{Baseline Answer Selection 4 -- Fine-tuning on DQA with query rewriting}
This experiment resembles Baseline $3$, but used the same query rewriting as in Baseline $2$ to the train and the hold-out set. 

\krimp
\subsubsection{Baseline Answer Selection 5 - Fine-tuning on SQuAD2.0 \& DQA}
This baseline resembles Baseline 1, but now we added our data to the existing SQuAD$2.0$ data set while fine-tuning the BERT Large model. We did this since our DQA dataset is not very large. We expected an improvement in performance when we enhanced our data with more data points. We shuffled the training input randomly. We evaluated on the DQA hold-out set.

\krimp
\subsubsection{Baseline Answer Selection 6 - Fine-tuning on SQuAD$2.0$ \& DQA with query rewriting}
This baseline resembles Baseline $5$, but we performed the same query rewriting to DQA part of the train set and to the DQA hold-out set as in Baselines $2$ and $4$.

\krimp
\subsection{Results for Answer Selection}
Table \ref{tab:results_answer_selection_v2} shows the results for the answer selection experiments. Fine-tuning BERT on SQuAD$2.0$ and the DQA data significantly outperforms the other baselines. These results look promising but reveal an interesting new problem to work on as the scores are significantly lower than we are used to from the typical QA task leader boards such as the SQuAD$2.0$ challenge. It is interesting to see that query rewriting is not beneficial. We assume that our approach may have been too simplistic. We would like to experiment with different types of query rewriting in future work (e.g., \cite{zhang2007query,grbovic2015context}).

\subsection{Answering RQ3}
We have shown that the initial baseline models perform reasonably well on our new document-centered domain. For the answer selection task, it is beneficial to add data from the Wikipedia domain (SQuAD2.0) during training. This improves the results, but also shows that document-centered assistance is a very different novel domain. While our initial experimental results are promising, there is still plenty of opportunity to improve the models in future work, for example by increasing the dataset size. We also expect improvements if we would train BERT on data similar to the DQA data. As BERT has been trained on the Wikipedia domain -- the same domain as the SQuAD2.0 data -- BERT could `memorize' certain parts of the data during training, which could give an advantage when fine-tuning on the SQuAD2.0 data. DQA does not have this advantage. In some specific scenarios, using the meta-structure of the document might help to improve results. However, we consider not relying on this structure as the preferred option since this allows us to generalize quickly over a wide variety of documents.

\begin{table}
\caption{Results answer selection. All models fine-tuned BERT Large and were evaluated on the DQA hold-out set. AS means Answer Selection. AS 5 significantly outperforms the other baselines (Wilcoxon Signed-rank, $p < 0.001$).}
\label{tab:results_answer_selection_v2}
\begin{tabular}{llll}
\toprule
	\textbf{Baseline} & \textbf{Training source} &  \textbf{F1} & \textbf{EM} \\ \midrule
	AS 1 & SQuAD2.0 &  27.24 & 13.21 \\
	AS 2 & SQuAD2.0 & & \\ 
	& with Eval Query rewriting & 26.79 & 13.09 \\
	AS 3 & DQA & 38.84 & 18.93 \\
	AS 4 & DQA with Query rewriting & 36.73 & 17.83 \\
	AS 5 & SQuAD2.0 + DQA & \textbf{41.02**} & \textbf{20.30**} \\
	AS 6 & SQuAD2.0 + DQA & & \\
	& with Query rewriting & 37.28 & 18.52 \\
	\bottomrule	
\end{tabular}
\vspace*{1em}
\end{table}

%% file: 06-discussion_conclusion.tex
%% !TEX root = ./main.tex
\section{Conclusions and future work}
In this paper, we explored the novel domain of document-centered digital assistance. We focused on a consumption scenario, in which individuals are a (co-)owner of a document.
Through a survey, we identified a set of primary capabilities people expect from a digital assistant in a document-centered scenario, as well as a large set of questions that gave us insight into the types of queries that people might pose about a document when they have an approximate or good idea what the document is about.
Our explorations shed light on the hierarchy of questions that might be posed, and demonstrate that the types of questions people ask in a document-centered scenario are different from the factoid questions in conventional QA datasets. We show that state-of-the-art QA models can be fine-tuned to perform with reasonable accuracy on the new DQA data. Yet, it has proven to be an unsolved task, which makes this a fertile area for future work. 
This research opens a new direction for digital assistance. Avenues for future work include deeper explorations of query rewriting to better tailor document-centered questions to conventional QA systems, and also exploring ways to scale up the data to a much larger and broader range of documents.

%% file: 07-AppendixA.tex
%% !TEX root = ./main.tex

\section{Data collection - Answers to questions}
\label{sec:appendix_A}

In this appendix we show additional screen shots of our answer selection procedure in Step 2 of this research, which was discussed in Section \ref{sec:data_collection}.

\begin{figure*}
	\centering
	\includegraphics[width=0.85\textwidth]{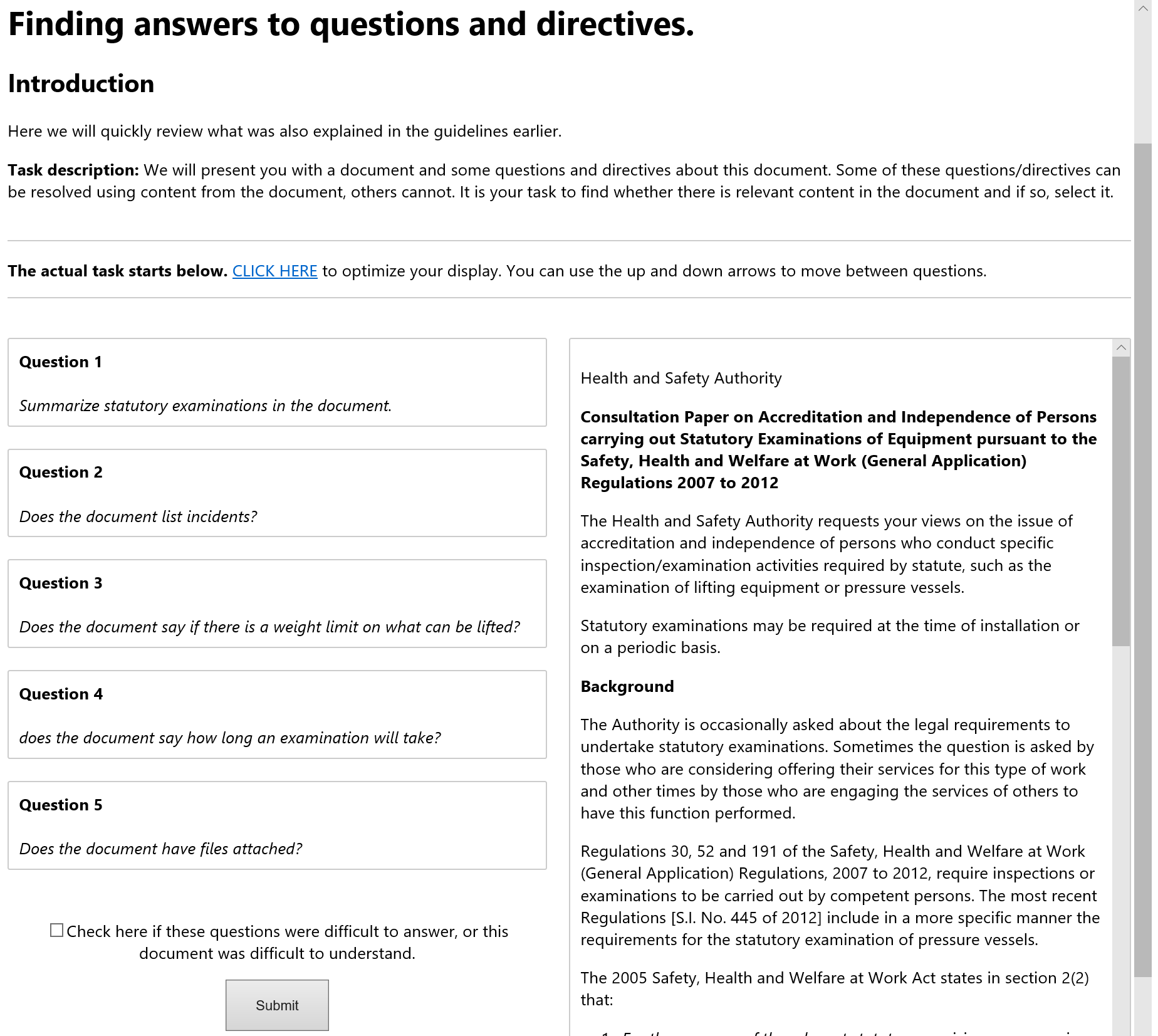}
	\caption{Question answering data collection overview.}
	\label{fig:sample_QA_overview}
\end{figure*}

\begin{figure*}
	\centering
	\includegraphics[width=0.37\textwidth]{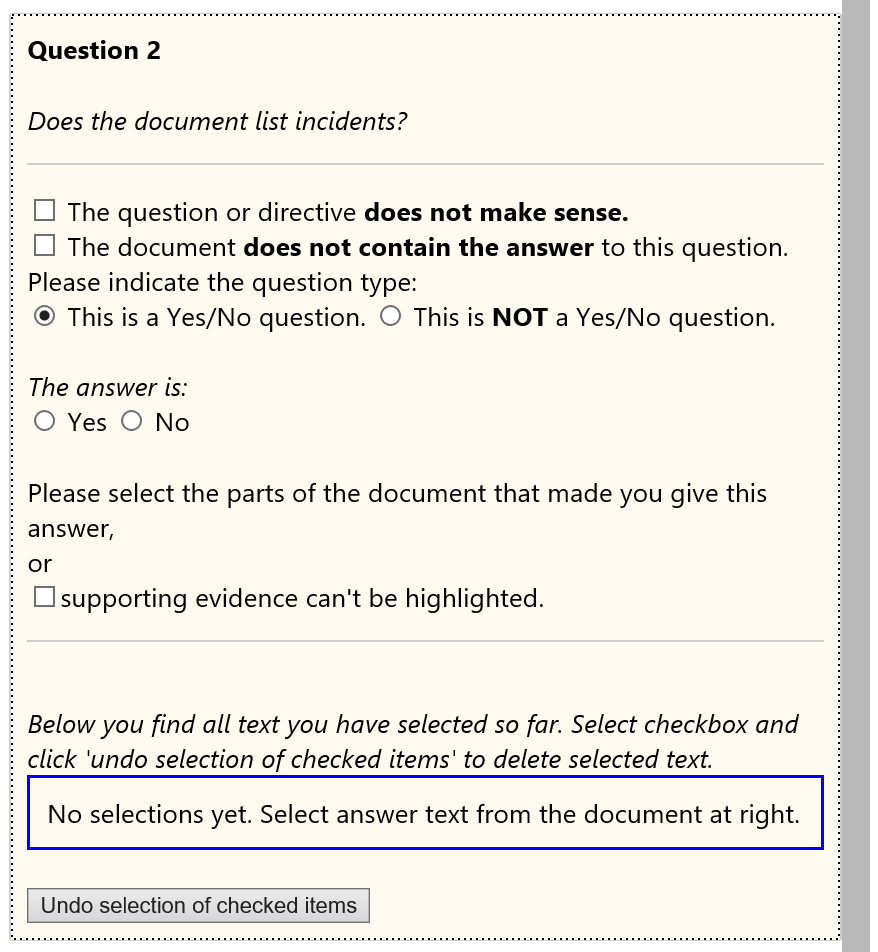}
	\caption{Question answering data collection yes/no expansion.}
	\label{fig:sample_QA_yesno}
\end{figure*}

%% file: main.bbl

\begin{thebibliography}{41}

%%% the Web address.

\ifx \showCODEN    \undefined \def \showCODEN     #1{\unskip}     \fi
\ifx \showDOI      \undefined \def \showDOI       #1{#1}\fi
\ifx \showISBNx    \undefined \def \showISBNx     #1{\unskip}     \fi
\ifx \showISBNxiii \undefined \def \showISBNxiii  #1{\unskip}     \fi
\ifx \showISSN     \undefined \def \showISSN      #1{\unskip}     \fi
\ifx \showLCCN     \undefined \def \showLCCN      #1{\unskip}     \fi
\ifx \shownote     \undefined \def \shownote      #1{#1}          \fi
\ifx \showarticletitle \undefined \def \showarticletitle #1{#1}   \fi
\ifx \showURL      \undefined \def \showURL       {\relax}        \fi
% invisible to TeX
\providecommand\bibfield[2]{#2}
\providecommand\bibinfo[2]{#2}
\providecommand\natexlab[1]{#1}
\providecommand\showeprint[2][]{arXiv:#2}

\bibitem[\protect\citeauthoryear{Ai, Dumais, Craswell, and Liebling}{Ai
  et~al\mbox{.}}{2017}]%
        {ai2017characterizing}
\bibfield{author}{\bibinfo{person}{Qingyao Ai}, \bibinfo{person}{Susan~T
  Dumais}, \bibinfo{person}{Nick Craswell}, {and} \bibinfo{person}{Dan
  Liebling}.} \bibinfo{year}{2017}\natexlab{}.
\newblock \showarticletitle{Characterizing email search using large-scale
  behavioral logs and surveys}. In \bibinfo{booktitle}{\emph{Proceedings of the
  26th International Conference on World Wide Web}}.
  \bibinfo{publisher}{International World Wide Web Conferences Steering
  Committee}, \bibinfo{address}{Republic and Canton of Geneva, Switzerland},
  \bibinfo{pages}{1511--1520}.
\newblock


\bibitem[\protect\citeauthoryear{Ashok, Borodin, Puzis, and Ramakrishnan}{Ashok
  et~al\mbox{.}}{2015}]%
        {ashok2015capti}
\bibfield{author}{\bibinfo{person}{Vikas Ashok}, \bibinfo{person}{Yevgen
  Borodin}, \bibinfo{person}{Yury Puzis}, {and} \bibinfo{person}{IV
  Ramakrishnan}.} \bibinfo{year}{2015}\natexlab{}.
\newblock \showarticletitle{Capti-speak: a speech-enabled web screen reader}.
  In \bibinfo{booktitle}{\emph{Proceedings of the 12th Web for All
  Conference}}. ACM, \bibinfo{pages}{22}.
\newblock


\bibitem[\protect\citeauthoryear{Bordes, Usunier, Chopra, and Weston}{Bordes
  et~al\mbox{.}}{2015}]%
        {bordes2015large}
\bibfield{author}{\bibinfo{person}{Antoine Bordes}, \bibinfo{person}{Nicolas
  Usunier}, \bibinfo{person}{Sumit Chopra}, {and} \bibinfo{person}{Jason
  Weston}.} \bibinfo{year}{2015}\natexlab{}.
\newblock \showarticletitle{Large-scale simple question answering with memory
  networks}.
\newblock \bibinfo{journal}{\emph{arXiv preprint arXiv:1506.02075}}
  (\bibinfo{year}{2015}).
\newblock


\bibitem[\protect\citeauthoryear{Bota, Fourney, Dumais, Religa, and
  Rounthwaite}{Bota et~al\mbox{.}}{2018}]%
        {bota2018characterizing}
\bibfield{author}{\bibinfo{person}{Horatiu Bota}, \bibinfo{person}{Adam
  Fourney}, \bibinfo{person}{Susan~T Dumais}, \bibinfo{person}{Tomasz~L
  Religa}, {and} \bibinfo{person}{Robert Rounthwaite}.}
  \bibinfo{year}{2018}\natexlab{}.
\newblock \showarticletitle{Characterizing Search Behavior in Productivity
  Software}. In \bibinfo{booktitle}{\emph{Proceedings of the 2018 Conference on
  Human Information Interaction \& Retrieval}}. ACM, \bibinfo{pages}{160--169}.
\newblock


\bibitem[\protect\citeauthoryear{Chen, Fisch, Weston, and Bordes}{Chen
  et~al\mbox{.}}{2017}]%
        {chen2017reading}
\bibfield{author}{\bibinfo{person}{Danqi Chen}, \bibinfo{person}{Adam Fisch},
  \bibinfo{person}{Jason Weston}, {and} \bibinfo{person}{Antoine Bordes}.}
  \bibinfo{year}{2017}\natexlab{}.
\newblock \showarticletitle{Reading wikipedia to answer open-domain questions}.
\newblock \bibinfo{journal}{\emph{arXiv preprint arXiv:1704.00051}}
  (\bibinfo{year}{2017}).
\newblock


\bibitem[\protect\citeauthoryear{Daniel, Brink, Eloff, and Copley}{Daniel
  et~al\mbox{.}}{2019}]%
        {daniel2019towards}
\bibfield{author}{\bibinfo{person}{Jeanne~E Daniel}, \bibinfo{person}{Willie
  Brink}, \bibinfo{person}{Ryan Eloff}, {and} \bibinfo{person}{Charles
  Copley}.} \bibinfo{year}{2019}\natexlab{}.
\newblock \showarticletitle{Towards Automating Healthcare Question Answering in
  a Noisy Multilingual Low-Resource Setting}. In
  \bibinfo{booktitle}{\emph{Proceedings of the 57th Annual Meeting of the
  Association for Computational Linguistics}}. \bibinfo{publisher}{Association
  for Computational Linguistics}, \bibinfo{address}{Florence, Italy},
  \bibinfo{pages}{948--953}.
\newblock


\bibitem[\protect\citeauthoryear{Dehghani, Azarbonyad, Kamps, and
  de~Rijke}{Dehghani et~al\mbox{.}}{2019}]%
        {dehghani2019learning}
\bibfield{author}{\bibinfo{person}{Mostafa Dehghani}, \bibinfo{person}{Hosein
  Azarbonyad}, \bibinfo{person}{Jaap Kamps}, {and} \bibinfo{person}{Maarten de
  Rijke}.} \bibinfo{year}{2019}\natexlab{}.
\newblock \showarticletitle{Learning to Transform, Combine, and Reason in
  Open-Domain Question Answering}. In \bibinfo{booktitle}{\emph{Proceedings of
  the Twelfth ACM International Conference on Web Search and Data Mining}}
  \emph{(\bibinfo{series}{WSDM '19})}. \bibinfo{publisher}{ACM},
  \bibinfo{address}{New York, NY, USA}, \bibinfo{pages}{681--689}.
\newblock
\showISBNx{978-1-4503-5940-5}
\urldef\tempurl%
\url{https://doi.org/10.1145/3289600.3291012}
\showDOI{\tempurl}


\bibitem[\protect\citeauthoryear{Devlin, Chang, Lee, and Toutanova}{Devlin
  et~al\mbox{.}}{2018}]%
        {devlin2018bert}
\bibfield{author}{\bibinfo{person}{Jacob Devlin}, \bibinfo{person}{Ming-Wei
  Chang}, \bibinfo{person}{Kenton Lee}, {and} \bibinfo{person}{Kristina
  Toutanova}.} \bibinfo{year}{2018}\natexlab{}.
\newblock \showarticletitle{Bert: Pre-training of deep bidirectional
  transformers for language understanding}.
\newblock \bibinfo{journal}{\emph{arXiv preprint arXiv:1810.04805}}
  (\bibinfo{year}{2018}).
\newblock


\bibitem[\protect\citeauthoryear{Di~Geronimo, Husmann, and Norrie}{Di~Geronimo
  et~al\mbox{.}}{2016}]%
        {di2016surveying}
\bibfield{author}{\bibinfo{person}{Linda Di~Geronimo}, \bibinfo{person}{Maria
  Husmann}, {and} \bibinfo{person}{Moira~C Norrie}.}
  \bibinfo{year}{2016}\natexlab{}.
\newblock \showarticletitle{Surveying personal device ecosystems with
  cross-device applications in mind}. In \bibinfo{booktitle}{\emph{Proceedings
  of the 5th ACM International Symposium on Pervasive Displays}}. ACM,
  \bibinfo{pages}{220--227}.
\newblock


\bibitem[\protect\citeauthoryear{Dunn, Sagun, Higgins, Guney, Cirik, and
  Cho}{Dunn et~al\mbox{.}}{2017}]%
        {dunn2017searchqa}
\bibfield{author}{\bibinfo{person}{Matthew Dunn}, \bibinfo{person}{Levent
  Sagun}, \bibinfo{person}{Mike Higgins}, \bibinfo{person}{V~Ugur Guney},
  \bibinfo{person}{Volkan Cirik}, {and} \bibinfo{person}{Kyunghyun Cho}.}
  \bibinfo{year}{2017}\natexlab{}.
\newblock \showarticletitle{Searchqa: A new q\&a dataset augmented with context
  from a search engine}.
\newblock \bibinfo{journal}{\emph{arXiv preprint arXiv:1704.05179}}
  (\bibinfo{year}{2017}).
\newblock


\bibitem[\protect\citeauthoryear{Fourney and Dumais}{Fourney and
  Dumais}{2016}]%
        {fourney2016automatic}
\bibfield{author}{\bibinfo{person}{Adam Fourney} {and} \bibinfo{person}{Susan~T
  Dumais}.} \bibinfo{year}{2016}\natexlab{}.
\newblock \showarticletitle{Automatic identification and contextual
  reformulation of implicit system-related queries}. In
  \bibinfo{booktitle}{\emph{Proceedings of the 39th International ACM SIGIR
  conference on Research and Development in Information Retrieval}}. ACM,
  \bibinfo{pages}{761--764}.
\newblock


\bibitem[\protect\citeauthoryear{Gan and Ng}{Gan and Ng}{2019}]%
        {gan2019improving}
\bibfield{author}{\bibinfo{person}{Wee~Chung Gan} {and}
  \bibinfo{person}{Hwee~Tou Ng}.} \bibinfo{year}{2019}\natexlab{}.
\newblock \showarticletitle{Improving the Robustness of Question Answering
  Systems to Question Paraphrasing}. In \bibinfo{booktitle}{\emph{Proceedings
  of the 57th Annual Meeting of the Association for Computational
  Linguistics}}. \bibinfo{publisher}{Association for Computational
  Linguistics}, \bibinfo{address}{Florence, Italy},
  \bibinfo{pages}{6065--6075}.
\newblock


\bibitem[\protect\citeauthoryear{Grbovic, Djuric, Radosavljevic, Silvestri, and
  Bhamidipati}{Grbovic et~al\mbox{.}}{2015}]%
        {grbovic2015context}
\bibfield{author}{\bibinfo{person}{Mihajlo Grbovic}, \bibinfo{person}{Nemanja
  Djuric}, \bibinfo{person}{Vladan Radosavljevic}, \bibinfo{person}{Fabrizio
  Silvestri}, {and} \bibinfo{person}{Narayan Bhamidipati}.}
  \bibinfo{year}{2015}\natexlab{}.
\newblock \showarticletitle{Context-and content-aware embeddings for query
  rewriting in sponsored search}. In \bibinfo{booktitle}{\emph{Proceedings of
  the 38th international ACM SIGIR conference on research and development in
  information retrieval}}. ACM, \bibinfo{pages}{383--392}.
\newblock


\bibitem[\protect\citeauthoryear{Jokela, Ojala, and Olsson}{Jokela
  et~al\mbox{.}}{2015}]%
        {jokela2015diary}
\bibfield{author}{\bibinfo{person}{Tero Jokela}, \bibinfo{person}{Jarno Ojala},
  {and} \bibinfo{person}{Thomas Olsson}.} \bibinfo{year}{2015}\natexlab{}.
\newblock \showarticletitle{A diary study on combining multiple information
  devices in everyday activities and tasks}. In
  \bibinfo{booktitle}{\emph{Proceedings of the 33rd Annual ACM Conference on
  Human Factors in Computing Systems}}. ACM, \bibinfo{pages}{3903--3912}.
\newblock


\bibitem[\protect\citeauthoryear{Joshi, Choi, Weld, and Zettlemoyer}{Joshi
  et~al\mbox{.}}{2017}]%
        {joshi2017triviaqa}
\bibfield{author}{\bibinfo{person}{Mandar Joshi}, \bibinfo{person}{Eunsol
  Choi}, \bibinfo{person}{Daniel~S Weld}, {and} \bibinfo{person}{Luke
  Zettlemoyer}.} \bibinfo{year}{2017}\natexlab{}.
\newblock \showarticletitle{Triviaqa: A large scale distantly supervised
  challenge dataset for reading comprehension}.
\newblock \bibinfo{journal}{\emph{arXiv preprint arXiv:1705.03551}}
  (\bibinfo{year}{2017}).
\newblock


\bibitem[\protect\citeauthoryear{Karlson, Iqbal, Meyers, Ramos, Lee, and
  Tang}{Karlson et~al\mbox{.}}{2010}]%
        {karlson2010mobile}
\bibfield{author}{\bibinfo{person}{Amy~K Karlson}, \bibinfo{person}{Shamsi~T
  Iqbal}, \bibinfo{person}{Brian Meyers}, \bibinfo{person}{Gonzalo Ramos},
  \bibinfo{person}{Kathy Lee}, {and} \bibinfo{person}{John~C Tang}.}
  \bibinfo{year}{2010}\natexlab{}.
\newblock \showarticletitle{Mobile taskflow in context: a screenshot study of
  smartphone usage}. In \bibinfo{booktitle}{\emph{Proceedings of the SIGCHI
  Conference on Human Factors in Computing Systems}}. ACM,
  \bibinfo{pages}{2009--2018}.
\newblock


\bibitem[\protect\citeauthoryear{Ko{\v{c}}isk{\`y}, Schwarz, Blunsom, Dyer,
  Hermann, Melis, and Grefenstette}{Ko{\v{c}}isk{\`y} et~al\mbox{.}}{2018}]%
        {kovcisky2018narrativeqa}
\bibfield{author}{\bibinfo{person}{Tom{\'a}{\v{s}} Ko{\v{c}}isk{\`y}},
  \bibinfo{person}{Jonathan Schwarz}, \bibinfo{person}{Phil Blunsom},
  \bibinfo{person}{Chris Dyer}, \bibinfo{person}{Karl~Moritz Hermann},
  \bibinfo{person}{G{\'a}bor Melis}, {and} \bibinfo{person}{Edward
  Grefenstette}.} \bibinfo{year}{2018}\natexlab{}.
\newblock \showarticletitle{The narrativeqa reading comprehension challenge}.
\newblock \bibinfo{journal}{\emph{Transactions of the Association for
  Computational Linguistics}}  \bibinfo{volume}{6} (\bibinfo{year}{2018}),
  \bibinfo{pages}{317--328}.
\newblock


\bibitem[\protect\citeauthoryear{Kratzwald, Eigenmann, and
  Feuerriegel}{Kratzwald et~al\mbox{.}}{2019}]%
        {kratzwald2019rankqa}
\bibfield{author}{\bibinfo{person}{Bernhard Kratzwald}, \bibinfo{person}{Anna
  Eigenmann}, {and} \bibinfo{person}{Stefan Feuerriegel}.}
  \bibinfo{year}{2019}\natexlab{}.
\newblock \showarticletitle{RankQA: Neural Question Answering with Answer
  Re-Ranking}.
\newblock \bibinfo{journal}{\emph{arXiv preprint arXiv:1906.03008}}
  (\bibinfo{year}{2019}).
\newblock


\bibitem[\protect\citeauthoryear{Kwiatkowski, Palomaki, Redfield, Collins,
  Parikh, Alberti, Epstein, Polosukhin, Devlin, Lee, et~al\mbox{.}}{Kwiatkowski
  et~al\mbox{.}}{2019}]%
        {kwiatkowski2019natural}
\bibfield{author}{\bibinfo{person}{Tom Kwiatkowski},
  \bibinfo{person}{Jennimaria Palomaki}, \bibinfo{person}{Olivia Redfield},
  \bibinfo{person}{Michael Collins}, \bibinfo{person}{Ankur Parikh},
  \bibinfo{person}{Chris Alberti}, \bibinfo{person}{Danielle Epstein},
  \bibinfo{person}{Illia Polosukhin}, \bibinfo{person}{Jacob Devlin},
  \bibinfo{person}{Kenton Lee}, {et~al\mbox{.}}}
  \bibinfo{year}{2019}\natexlab{}.
\newblock \showarticletitle{Natural questions: a benchmark for question
  answering research}.
\newblock \bibinfo{journal}{\emph{Transactions of the Association for
  Computational Linguistics}}  \bibinfo{volume}{7} (\bibinfo{year}{2019}),
  \bibinfo{pages}{453--466}.
\newblock


\bibitem[\protect\citeauthoryear{Lewis, Denoyer, and Riedel}{Lewis
  et~al\mbox{.}}{2019}]%
        {lewis2019unsupervised}
\bibfield{author}{\bibinfo{person}{Patrick Lewis}, \bibinfo{person}{Ludovic
  Denoyer}, {and} \bibinfo{person}{Sebastian Riedel}.}
  \bibinfo{year}{2019}\natexlab{}.
\newblock \showarticletitle{Unsupervised Question Answering by Cloze
  Translation}.
\newblock \bibinfo{journal}{\emph{arXiv preprint arXiv:1906.04980}}
  (\bibinfo{year}{2019}).
\newblock


\bibitem[\protect\citeauthoryear{Li}{Li}{2019}]%
        {li2019probabilistic}
\bibfield{author}{\bibinfo{person}{Yuan Li}.} \bibinfo{year}{2019}\natexlab{}.
\newblock \emph{\bibinfo{title}{Probabilistic models for aggregating
  crowdsourced annotations}}.
\newblock \bibinfo{thesistype}{Ph.D. Dissertation}.
\newblock


\bibitem[\protect\citeauthoryear{Lin}{Lin}{2004}]%
        {lin2004rouge}
\bibfield{author}{\bibinfo{person}{Chin-Yew Lin}.}
  \bibinfo{year}{2004}\natexlab{}.
\newblock \showarticletitle{{ROUGE}: A Package for Automatic Evaluation of
  Summaries}. In \bibinfo{booktitle}{\emph{Text Summarization Branches Out}}.
  \bibinfo{publisher}{Association for Computational Linguistics},
  \bibinfo{address}{Barcelona, Spain}, \bibinfo{pages}{74--81}.
\newblock


\bibitem[\protect\citeauthoryear{Lo and Green}{Lo and Green}{2013}]%
        {lo2013development}
\bibfield{author}{\bibinfo{person}{Victor Ei-Wen Lo} {and}
  \bibinfo{person}{Paul~A Green}.} \bibinfo{year}{2013}\natexlab{}.
\newblock \showarticletitle{Development and evaluation of automotive speech
  interfaces: useful information from the human factors and the related
  literature}.
\newblock \bibinfo{journal}{\emph{International Journal of Vehicular
  Technology}}  \bibinfo{volume}{2013} (\bibinfo{year}{2013}).
\newblock


\bibitem[\protect\citeauthoryear{Martelaro, Teevan, and Iqbal}{Martelaro
  et~al\mbox{.}}{2019}]%
        {martelaro2019exploration}
\bibfield{author}{\bibinfo{person}{Nikolas Martelaro}, \bibinfo{person}{Jaime
  Teevan}, {and} \bibinfo{person}{Shamsi~T Iqbal}.}
  \bibinfo{year}{2019}\natexlab{}.
\newblock \showarticletitle{An Exploration of Speech-Based Productivity Support
  in the Car}. In \bibinfo{booktitle}{\emph{Proceedings of the 2019 CHI
  Conference on Human Factors in Computing Systems}}. ACM,
  \bibinfo{pages}{264}.
\newblock


\bibitem[\protect\citeauthoryear{Microsoft}{Microsoft}{2019}]%
        {voice_report}
\bibfield{author}{\bibinfo{person}{Microsoft}.}
  \bibinfo{year}{2019}\natexlab{}.
\newblock \bibinfo{title}{Voice report. From answers to action: customer
  adoption of voice technology and digital assistants}.
\newblock
  \bibinfo{howpublished}{\url{https://advertiseonbing-blob.azureedge.net/blob/bingads/media/insight/whitepapers/2019/04\%20apr/voice-report/bingads_2019_voicereport.pdf}}.
\newblock
\newblock
\shownote{Accessed: 2019-12-04.}


\bibitem[\protect\citeauthoryear{Nguyen, Rosenberg, Song, Gao, Tiwary,
  Majumder, and Deng}{Nguyen et~al\mbox{.}}{2016}]%
        {nguyen2016ms}
\bibfield{author}{\bibinfo{person}{Tri Nguyen}, \bibinfo{person}{Mir
  Rosenberg}, \bibinfo{person}{Xia Song}, \bibinfo{person}{Jianfeng Gao},
  \bibinfo{person}{Saurabh Tiwary}, \bibinfo{person}{Rangan Majumder}, {and}
  \bibinfo{person}{Li Deng}.} \bibinfo{year}{2016}\natexlab{}.
\newblock \showarticletitle{MS MARCO: A Human-Generated MAchine Reading
  COmprehension Dataset}.
\newblock  (\bibinfo{year}{2016}).
\newblock


\bibitem[\protect\citeauthoryear{Rajpurkar, Jia, and Liang}{Rajpurkar
  et~al\mbox{.}}{2018}]%
        {rajpurkar2018know}
\bibfield{author}{\bibinfo{person}{Pranav Rajpurkar}, \bibinfo{person}{Robin
  Jia}, {and} \bibinfo{person}{Percy Liang}.} \bibinfo{year}{2018}\natexlab{}.
\newblock \showarticletitle{Know What You Don't Know: Unanswerable Questions
  for SQuAD}.
\newblock \bibinfo{journal}{\emph{arXiv preprint arXiv:1806.03822}}
  (\bibinfo{year}{2018}).
\newblock


\bibitem[\protect\citeauthoryear{Rajpurkar, Zhang, Lopyrev, and
  Liang}{Rajpurkar et~al\mbox{.}}{2016}]%
        {rajpurkar2016squad}
\bibfield{author}{\bibinfo{person}{Pranav Rajpurkar}, \bibinfo{person}{Jian
  Zhang}, \bibinfo{person}{Konstantin Lopyrev}, {and} \bibinfo{person}{Percy
  Liang}.} \bibinfo{year}{2016}\natexlab{}.
\newblock \showarticletitle{Squad: 100,000+ questions for machine comprehension
  of text}.
\newblock \bibinfo{journal}{\emph{arXiv preprint arXiv:1606.05250}}
  (\bibinfo{year}{2016}).
\newblock


\bibitem[\protect\citeauthoryear{Reddy, Chen, and Manning}{Reddy
  et~al\mbox{.}}{2019}]%
        {reddy2019coqa}
\bibfield{author}{\bibinfo{person}{Siva Reddy}, \bibinfo{person}{Danqi Chen},
  {and} \bibinfo{person}{Christopher~D Manning}.}
  \bibinfo{year}{2019}\natexlab{}.
\newblock \showarticletitle{Coqa: A conversational question answering
  challenge}.
\newblock \bibinfo{journal}{\emph{Transactions of the Association for
  Computational Linguistics}}  \bibinfo{volume}{7} (\bibinfo{year}{2019}),
  \bibinfo{pages}{249--266}.
\newblock


\bibitem[\protect\citeauthoryear{Robertson, Zaragoza, et~al\mbox{.}}{Robertson
  et~al\mbox{.}}{2009}]%
        {robertson2009probabilistic}
\bibfield{author}{\bibinfo{person}{Stephen Robertson}, \bibinfo{person}{Hugo
  Zaragoza}, {et~al\mbox{.}}} \bibinfo{year}{2009}\natexlab{}.
\newblock \showarticletitle{The probabilistic relevance framework: BM25 and
  beyond}.
\newblock \bibinfo{journal}{\emph{Foundations and Trends{\textregistered} in
  Information Retrieval}} \bibinfo{volume}{3}, \bibinfo{number}{4}
  (\bibinfo{year}{2009}), \bibinfo{pages}{333--389}.
\newblock


\bibitem[\protect\citeauthoryear{Sankar and Ravi}{Sankar and Ravi}{2018}]%
        {sankar2018modeling}
\bibfield{author}{\bibinfo{person}{Chinnadhurai Sankar} {and}
  \bibinfo{person}{Sujith Ravi}.} \bibinfo{year}{2018}\natexlab{}.
\newblock \showarticletitle{Modeling non-goal oriented dialog with discrete
  attributes}. In \bibinfo{booktitle}{\emph{NeurIPS Workshop on Conversational
  AI:“Today’s Practice and Tomorrow‘s Potential}}.
\newblock


\bibitem[\protect\citeauthoryear{Trischler, Wang, Yuan, Harris, Sordoni,
  Bachman, and Suleman}{Trischler et~al\mbox{.}}{2016}]%
        {trischler2016newsqa}
\bibfield{author}{\bibinfo{person}{Adam Trischler}, \bibinfo{person}{Tong
  Wang}, \bibinfo{person}{Xingdi Yuan}, \bibinfo{person}{Justin Harris},
  \bibinfo{person}{Alessandro Sordoni}, \bibinfo{person}{Philip Bachman}, {and}
  \bibinfo{person}{Kaheer Suleman}.} \bibinfo{year}{2016}\natexlab{}.
\newblock \showarticletitle{Newsqa: A machine comprehension dataset}.
\newblock \bibinfo{journal}{\emph{arXiv preprint arXiv:1611.09830}}
  (\bibinfo{year}{2016}).
\newblock


\bibitem[\protect\citeauthoryear{Tsai}{Tsai}{2018}]%
        {ai_chatbots_intelligent_assistants_workplace}
\bibfield{author}{\bibinfo{person}{Peter Tsai}.}
  \bibinfo{year}{2018}\natexlab{}.
\newblock \bibinfo{title}{Data snapshot: AI Chatbots and Intelligent Assistants
  in the Workplace}.
\newblock
  \bibinfo{howpublished}{\url{https://community.spiceworks.com/blog/2964-data-snapshot-ai-chatbots-and-intelligent-assistants-in-the-workplace}}.
\newblock
\newblock
\shownote{Accessed: 2019-10-04.}


\bibitem[\protect\citeauthoryear{Vtyurina, Fourney, Morris, Findlater, and
  White}{Vtyurina et~al\mbox{.}}{2019}]%
        {vtyurina2019verse}
\bibfield{author}{\bibinfo{person}{Alexandra Vtyurina}, \bibinfo{person}{Adam
  Fourney}, \bibinfo{person}{Meredith~Ringel Morris}, \bibinfo{person}{Leah
  Findlater}, {and} \bibinfo{person}{Ryen White}.}
  \bibinfo{year}{2019}\natexlab{}.
\newblock \showarticletitle{VERSE: Bridging Screen Readers and Voice Assistants
  for Enhanced Eyes-Free Web Search}. In \bibinfo{booktitle}{\emph{Proceedings
  of the 21st International ACM SIGACCESS Conference on Computers and
  Accessibility. ACM}}.
\newblock


\bibitem[\protect\citeauthoryear{Williams, Kaur, Iqbal, White, Teevan, and
  Fourney}{Williams et~al\mbox{.}}{2019}]%
        {williams2019mercury}
\bibfield{author}{\bibinfo{person}{Alex~C Williams},
  \bibinfo{person}{Harmanpreet Kaur}, \bibinfo{person}{Shamsi Iqbal},
  \bibinfo{person}{Ryen~W White}, \bibinfo{person}{Jaime Teevan}, {and}
  \bibinfo{person}{Adam Fourney}.} \bibinfo{year}{2019}\natexlab{}.
\newblock \showarticletitle{Mercury: Empowering Programmers' Mobile Work
  Practices with Microproductivity}. In \bibinfo{booktitle}{\emph{Proceedings
  of the 32nd Annual ACM Symposium on User Interface Software and Technology.
  ACM Press, New Orleans, Louisana, USA. https://doi.
  org/10.1145/3332165.3347932}}.
\newblock


\bibitem[\protect\citeauthoryear{Wolf, Debut, Sanh, Chaumond, Delangue, Moi,
  Cistac, Rault, Louf, Funtowicz, and Brew}{Wolf et~al\mbox{.}}{2019}]%
        {Wolf2019HuggingFacesTS}
\bibfield{author}{\bibinfo{person}{Thomas Wolf}, \bibinfo{person}{Lysandre
  Debut}, \bibinfo{person}{Victor Sanh}, \bibinfo{person}{Julien Chaumond},
  \bibinfo{person}{Clement Delangue}, \bibinfo{person}{Anthony Moi},
  \bibinfo{person}{Pierric Cistac}, \bibinfo{person}{Tim Rault},
  \bibinfo{person}{R'emi Louf}, \bibinfo{person}{Morgan Funtowicz}, {and}
  \bibinfo{person}{Jamie Brew}.} \bibinfo{year}{2019}\natexlab{}.
\newblock \showarticletitle{HuggingFace's Transformers: State-of-the-art
  Natural Language Processing}.
\newblock \bibinfo{journal}{\emph{ArXiv}}  \bibinfo{volume}{abs/1910.03771}
  (\bibinfo{year}{2019}).
\newblock


\bibitem[\protect\citeauthoryear{Yan and Zhao}{Yan and Zhao}{2018}]%
        {yan2018coupled}
\bibfield{author}{\bibinfo{person}{Rui Yan} {and} \bibinfo{person}{Dongyan
  Zhao}.} \bibinfo{year}{2018}\natexlab{}.
\newblock \showarticletitle{Coupled context modeling for deep chit-chat:
  towards conversations between human and computer}. In
  \bibinfo{booktitle}{\emph{Proceedings of the 24th ACM SIGKDD International
  Conference on Knowledge Discovery \& Data Mining}}. ACM,
  \bibinfo{pages}{2574--2583}.
\newblock


\bibitem[\protect\citeauthoryear{Yang, Xie, Tan, Xiong, Li, and Lin}{Yang
  et~al\mbox{.}}{2019}]%
        {yang2019data}
\bibfield{author}{\bibinfo{person}{Wei Yang}, \bibinfo{person}{Yuqing Xie},
  \bibinfo{person}{Luchen Tan}, \bibinfo{person}{Kun Xiong},
  \bibinfo{person}{Ming Li}, {and} \bibinfo{person}{Jimmy Lin}.}
  \bibinfo{year}{2019}\natexlab{}.
\newblock \showarticletitle{Data Augmentation for BERT Fine-Tuning in
  Open-Domain Question Answering}.
\newblock \bibinfo{journal}{\emph{arXiv preprint arXiv:1904.06652}}
  (\bibinfo{year}{2019}).
\newblock


\bibitem[\protect\citeauthoryear{Yang, Yih, and Meek}{Yang
  et~al\mbox{.}}{2015}]%
        {yang2015wikiqa}
\bibfield{author}{\bibinfo{person}{Yi Yang}, \bibinfo{person}{Wen-tau Yih},
  {and} \bibinfo{person}{Christopher Meek}.} \bibinfo{year}{2015}\natexlab{}.
\newblock \showarticletitle{Wikiqa: A challenge dataset for open-domain
  question answering}. In \bibinfo{booktitle}{\emph{Proceedings of the 2015
  Conference on Empirical Methods in Natural Language Processing}}.
  \bibinfo{publisher}{Association for Computational Linguistics},
  \bibinfo{address}{Lisbon, Portugal}, \bibinfo{pages}{2013--2018}.
\newblock


\bibitem[\protect\citeauthoryear{Yang, Qi, Zhang, Bengio, Cohen, Salakhutdinov,
  and Manning}{Yang et~al\mbox{.}}{2018}]%
        {yang2018hotpotqa}
\bibfield{author}{\bibinfo{person}{Zhilin Yang}, \bibinfo{person}{Peng Qi},
  \bibinfo{person}{Saizheng Zhang}, \bibinfo{person}{Yoshua Bengio},
  \bibinfo{person}{William~W Cohen}, \bibinfo{person}{Ruslan Salakhutdinov},
  {and} \bibinfo{person}{Christopher~D Manning}.}
  \bibinfo{year}{2018}\natexlab{}.
\newblock \showarticletitle{Hotpotqa: A dataset for diverse, explainable
  multi-hop question answering}.
\newblock \bibinfo{journal}{\emph{arXiv preprint arXiv:1809.09600}}
  (\bibinfo{year}{2018}).
\newblock


\bibitem[\protect\citeauthoryear{Zhang, He, Rey, and Jones}{Zhang
  et~al\mbox{.}}{2007}]%
        {zhang2007query}
\bibfield{author}{\bibinfo{person}{Wei~Vivian Zhang}, \bibinfo{person}{Xiaofei
  He}, \bibinfo{person}{Benjamin Rey}, {and} \bibinfo{person}{Rosie Jones}.}
  \bibinfo{year}{2007}\natexlab{}.
\newblock \showarticletitle{Query rewriting using active learning for sponsored
  search}. In \bibinfo{booktitle}{\emph{Proceedings of the 30th annual
  international ACM SIGIR conference on Research and development in information
  retrieval}}. ACM, \bibinfo{pages}{853--854}.
\newblock


\end{thebibliography}
